\documentclass[10pt,twocolumn,letterpaper]{article}

\usepackage{cvpr}      

\usepackage{amsmath,amsfonts,bm}

\def\eqref#1{equation~\ref{#1}}

\def\1{\bm{1}}

\def\vb{{\bm{b}}}

\def\vz{{\bm{z}}}

\def\mW{{\bm{W}}}

\DeclareMathAlphabet{\mathsfit}{\encodingdefault}{\sfdefault}{m}{sl}
\SetMathAlphabet{\mathsfit}{bold}{\encodingdefault}{\sfdefault}{bx}{n}

\usepackage{caption}
\usepackage{subcaption}
\usepackage{annotate-equations}
\usepackage{dsfont}
\usepackage[table]{xcolor}
\usepackage{tabularx}
\usepackage{multirow}
\usepackage[most, listings, theorems]{tcolorbox}

\newcommand{\stinformed}{\textit{Steering-informed Top-k}\xspace}
\newcommand{\steering}{\textit{Steering}\xspace}
\newcommand{\topk}{\textit{Top-k}\xspace}

\definecolor{darkgreen}{HTML}{138808}
\definecolor{green_drawio}{HTML}{82B366}
\definecolor{dark_green_drawio}{HTML}{557543}
\definecolor{dark_red_drawio}{HTML}{990000}
\definecolor{blue_drawio}{HTML}{6C8EBF}
\definecolor{orange_drawio}{HTML}{D79B00}
\definecolor{red_drawio}{HTML}{990000}
\definecolor{grey_drawio}{HTML}{303030}

\definecolor{cvprblue}{rgb}{0.21,0.49,0.74}
\usepackage[pagebackref,breaklinks,colorlinks,allcolors=cvprblue]{hyperref}

\def\paperID{21348} 
\def\confName{CVPR}
\def\confYear{2026}

\title{Language Models Can Explain Visual Features via Steering}

\author{\hspace{5pt}Javier Ferrando\thanks{Correspondance to \texttt{jferrandomonsonis@gmail.com}. This work is not related to the author’s position at Amazon.}
\hspace{1.5mm}
  Enrique Lopez-Cuena
\hspace{1.5mm}
  Pablo Agustin Martin-Torres \\
\hspace{1.5mm}
  \hspace{25pt}Daniel Hinjos
\hspace{1.5mm}
  Anna Arias-Duart
\hspace{1.5mm}
  Dario Garcia-Gasulla \\
\textbf{Barcelona Supercomputing Center}
}

\begin{document}
\maketitle
\begin{abstract}
Sparse Autoencoders uncover thousands of features in vision models, yet explaining these features without requiring human intervention remains an open challenge. While previous work has proposed generating correlation-based explanations based on top activating input examples, we present a fundamentally different alternative based on causal interventions. We leverage the structure of Vision-Language Models and \textit{steer} individual SAE features in the vision encoder after providing an empty image. Then, we prompt the language model to explain what it ``sees'', effectively eliciting the visual concept represented by each feature. Results show that \steering offers an scalable alternative that complements traditional approaches based on input examples, serving as a new axis for automated interpretability in vision models. Moreover, the quality of explanations improves consistently with the scale of the language model, highlighting our method as a promising direction for future research. Finally, we propose \stinformed, a hybrid approach that combines the strengths of causal interventions and input-based approaches to achieve state-of-the-art explanation quality without additional computational cost.\footnote{We make the codebase available at \url{https://github.com/HPAI-BSC/vision-interp}.}
\end{abstract}
\begin{figure*}[!t]
\begin{centering}
    \includegraphics[width=0.84\textwidth]{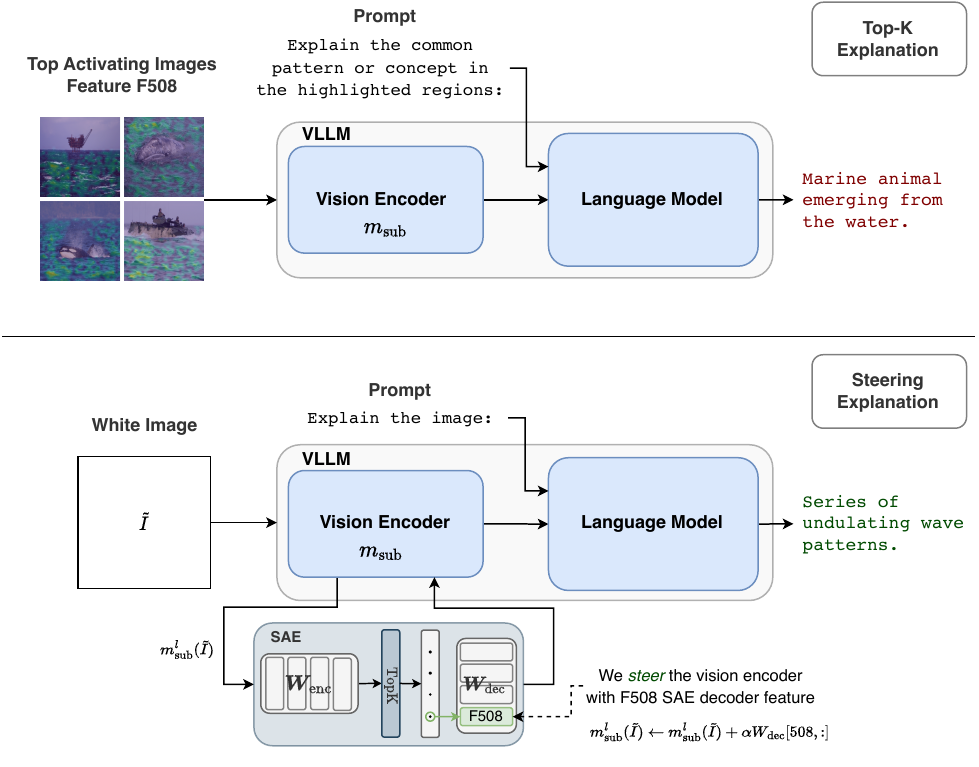}
    \caption{\textbf{Top}: A vision feature extracted with an SAE is explained based on top-activating images, which are passed to the VLM to generate an explanation based on correlated visual evidence. \textbf{Bottom}: We propose to automatically obtain explanations of SAE features by causally intervening (\textit{steering}) a vision encoder. The intervention is done after feeding it an information-devoid white image, effectively making the language model articulate what visual concept that feature represents.}
    \label{fig:figure_1}
\end{centering}
\end{figure*}    
\section{Introduction}
\label{sec:intro}

Understanding what features neural networks learn is a central goal in interpretability research~\citep{olah2020zoom}. Sparse Autoencoders (SAEs) have emerged as a promising unsupervised method for uncovering human-interpretable features from model representations~\citep{bricken2023monosemanticity,huben2024sparse}, particularly in large language models (LLMs). SAEs have recently been extended to vision models, revealing semantically meaningful concepts such as object categories, patterns, and textures~\citep{sae_vision,lim2025sparse}. However, as SAEs scale to uncover thousands of features, interpreting these poses significant challenges, necessitating the development of additional tools.

Recent work on automated interpretability aims to address this challenge by leveraging powerful language models as \textit{explainers} to generate descriptions of features learned by a \textit{subject} model~\citep{openai_neuron_nle,paulo2024automaticallyinterpretingmillionsfeatures}. When the subject is a vision model, the images that activate most each feature are analyzed by an explainer which looks for common patterns that may explain the target feature~\citep{xu2025decipheringfunctionsneuronsvisionlanguage,zhang2024largemultimodalmodelsinterpret}. This input-based strategy relies heavily on a predefined test set, is fundamentally correlation-based rather than causally grounded, and incurs significant computational cost.

An alternative line of work has explored \textit{self-explaining} approaches in language models, where the model itself is prompted to describe its representations~\citep{ghandeharioun2024patchscopes,chen2024selfie}. In this paper, we extend this paradigm to vision models by leveraging the structure of VLMs and perform causal interventions on SAE features\footnote{We refer to a `feature' as a direction in the model's representation space.} to describe them in natural language. By \textit{steering} the vision encoder’s residual stream with individual SAE features~—while feeding it an empty image—~we prompt the VLM to describe what visual concept that feature represents~(\Cref{fig:figure_1}). Experiments on Gemma 3 and Intern VL3 vision encoders show that \steering offers an scalable alternative that complements traditional approaches based on input examples,  overcoming some of the explanation biases these methods introduce, while surfacing lower-level features. Furthermore, scaling the language model consistently improves explanation quality, highlighting this causal, output-centric approach as a promising direction for automated interpretability.

Building on this idea, we also introduce a hybrid strategy~—\stinformed—~that combines the best of both approaches. We condition the VLM on the top activating images \emph{and} the causal intervention with the SAE feature, improving the quality of the generated explanations on four complementary metrics.
\section{Extracting Features}
Model neurons often exhibit polysemanticity, meaning they respond to seemingly unrelated concepts. One leading explanation for this phenomenon is \textit{superposition}, the idea that models learn to represent more concepts than they have neurons~\citep{arora-etal-2018-linear,elhage2022superposition}. Sparse Autoencoders (SAEs)~\citep{bricken2023monosemanticity} have emerged as an interpretability tool for finding interpretable and monosemantic features that are otherwise represented in superposition. SAEs achieve this by mapping model representations ${\vz \in \mathbb{R}^{d}}$ into a higher-dimensional latent space $\mathbb{R}^{d_{SAE}}$, while enforcing sparsity in the latent representation. In this work, we use TopK SAEs~\citep{gao2025scaling}, which apply the TopK activation function to enforce sparsity. The encoder first computes a sparse code using:
\begin{equation}\label{eq:topk_sae2}
    f(\vz) = \text{TopK}\bigl(\vz\mW_{\text{enc}} + \vb_{\text{enc}}\bigr),
\end{equation}
and the decoder reconstructs the original input from the sparse representation via
\begin{equation}\label{eq:topk_sae}
    \text{SAE}(\vz) =f(\vz)\mW_{\text{dec}} + \vb_{\text{dec}}.
\end{equation}
The encoder and decoder are parameterized by weight matrices and bias vectors $\mW_{\text{enc}}$, $\vb_{\text{enc}}$ and $\mW_{\text{dec}}$, $\vb_{\text{dec}}$ respectively. We refer to \textit{SAE feature activation} to a component in $f(\vz) \in \mathbb{R}^{\text{SAE}}$, while a \textit{SAE feature} denotes a row vector in the dictionary $\mW_{\text{dec}} \in \mathbb{R}^{d_{\text{SAE}}\times d_{\text{model}}}$. In this work, we train TopK SAEs with the latent space dimensionality $d_{\text{SAE}}=8,192$ on Imagenet dataset~\citep{imagenet}. We refer to Appendix~\ref{apx:sae_training} for further training details.
\section{Automatically Interpreting Features}
Following previous work in automated interpretability, we assume features can be explained by a sequence of words $\mathbf{e}$. We consider a \textit{subject} model $m_{\text{subj}}$ whose features we want to interpret, and an \textit{explainer} model $m_{\text{exp}}$ that generates the natural language explanations for these features.

\subsection{\topk Explanations}
The existing approach to generate explanations from vision model features~\citep{zhang2024largemultimodalmodelsinterpret,xu2025decipheringfunctionsneuronsvisionlanguage} assumes access to an evaluation set of images, $\mathcal{D}^{\text{eval}}$. Each image $I \in \mathcal{D}^{\text{eval}}$ is fed into the subject model $m_{\text{sub}}$, and the representations from the residual stream at a particular layer $l$ and position $j$, $m^{l,j}_{\text{sub}}(\cdot)$ are extracted; for brevity, we omit the layer index in what follows. Following~\Cref{eq:topk_sae2}, a SAE feature activation vector is obtained for each position $j$, $f(m_{\text{sub}}^{j}(I)) \in \mathbb{R}^{\text{SAE}}$. For each dimension $i \in \{1, \dots, d_{\text{SAE}}\}$, we compute an \textit{image activation score} by aggregating the individual position activations across the entire image:
\begin{equation}\label{eq:activation_score_image}
    S^{i,I} = g\left(f_i(m_{\text{sub}}(I))\right).
\end{equation}
Typically, the mean function (across positions) is used as $g(\cdot)$~\citep{zhang2024largemultimodalmodelsinterpret}. Then, we identify the top-$k$ images (with ${1 \leq k \leq |\mathcal{D}^{\text{eval}}|}$) that produce the highest \textit{image activation scores}. These images, denoted $\mathcal{T}^k_i = \{I^i_1,\hdots, I^i_k\}$, are selected such that their scores follow the descending order: ${S^{i,I^i_1} \geq S^{i,I^i_2} \geq \cdots \geq S^{i,I^i_{|{\mathcal{D}^{\text{eval}}}|}}}$. A natural language explanation $\mathbf{e}_i$ for the $i$-th feature is then generated by conditioning the explainer model on both a prompt $P$ and the selected top-$k$ images:
\begin{equation}
    \mathbf{e}_i \sim m_{\text{exp}}(\mathbf{e} \mid P,\mathcal{T}^k_i).
\end{equation}
Alternatively, the top-$k$ images can be modified to emphasize the regions where the feature is active. In our experiments, we explore two of such variants: `Masks', where all non-activating patches are occluded; and `Heatmaps', where activation intensity is overlaid to highlight the most responsive regions (see top activating images in~\Cref{fig:figure_1}).

\subsection{Proposed Approach}
Current VLMs align a visual encoder with a pre-trained language model backbone~\citep{bai2025qwen25vltechnicalreport,gemmateam2025gemma3technicalreport}, enabling natural image interpretation. We hypothesize that the language model can serve as an explainer for SAE features. We do so by causally intervening the vision encoder’s forward pass with each feature. We introduce two complementary methods for doing so.

\paragraph{\steering-based Explanations.}
In the basic setting, we prompt\footnote{The prompts used for each method can be found in Appendix~\ref{apx:prompts}.} the model to explain an empty image $\tilde{I}$, where by empty image we refer to a white image\footnote{We experimented with other types of images, such as black image, and random noise, obtaining similar results.} in which all pixels are assigned the same uniform white value, ensuring that it provides no meaningful visual signal to the encoder. Then, we intervene the forward pass by adding the SAE feature vector $\mW_{\text{dec}}[i,:]$ across all positions, effectively generating an explanation of the intervened feature. The process is formalized as follows:

\begin{equation}
\scalebox{0.9}{$
    \mathbf{e}_i \sim m_{\text{exp}}\left(\mathbf{e} \mid \eqnmarkbox[green_drawio]{prompt}{P},\eqnmarkbox[blue_drawio]{image}{\tilde{I}},\eqnmarkbox[orange_drawio]{intervention}{\text{do}(m_{\text{sub}}^l(\tilde{I}) \gets m_{\text{sub}}^l(\tilde{I}) + \alpha \mW_{\text{dec}}[i,:])}\right),
\annotate[yshift=0em]{above, left, label below}{prompt}{Prompt}
\annotate[yshift=0em]{above, right, label below}{image}{Empty image}
\annotate[yshift=-1em]{below, left, label below}{intervention}{Causal intervention with SAE feature}
$}\end{equation}
\vspace{4pt}

where we express the intervention using the do-operator~\citep{pearl_2009}, and $\alpha$ is a coefficient indicating the strength of the intervention.\footnote{In practice, we select the $\alpha$ coefficient on a validation set of 500 features.} By intervention we mean we substitute the model representation al layer $l$ across all positions by the SAE decoder vector. This method offers an efficient and scalable means of obtaining feature explanations, requiring a single forward-pass~(see Appendix \ref{apx:flops} for details). Unlike prior methods, it doesn't require an evaluation image set, simplifying the interpretability pipeline.

\paragraph{\stinformed Explanations.} Instead of only using a blank image, we apply the same causal intervention while conditioning on the top-$k$ images, $\mathcal{T}_i^k$—those that most strongly activate the $i$-th SAE feature. Intuitively, this focuses the explainer on the salient concept captured by the feature, enabling more targeted and meaningful interpretations. The process is defined as:

\begin{equation}
\scalebox{0.8}{$
    \mathbf{e}_i \sim m_{\text{exp}}\left(\mathbf{e} \mid \eqnmarkbox[green_drawio]{prompt}{P},\eqnmarkbox[red_drawio]{image}{\mathcal{T}^k_i},\eqnmarkbox[orange_drawio]{intervention}{\text{do}(m_{\text{sub}}^l(\mathcal{T}^k_i) \gets m_{\text{sub}}^l(\mathcal{T}^k_i) + \alpha \mW_{\text{dec}}[i,:])}\right).
\annotate[yshift=0em]{above, left, label below}{prompt}{Prompt}
\annotate[yshift=0em]{above, right, label below}{image}{Top-k images}
\annotate[yshift=-1em]{below, left, label below}{intervention}{Causal intervention with SAE feature}
$}\end{equation}
\vspace{6pt}
\section{Evaluating the Quality of the Explanations}\label{sec:evaluation_methods}

\subsection{Evaluation Metrics}\label{sec:evaluation_metrics}

To quantitatively assess the explanations, we adopt three complementary evaluation techniques. The first two are existing input-based evaluations relying on top-$k$ images~\citep{zhang2024largemultimodalmodelsinterpret,xu2025decipheringfunctionsneuronsvisionlanguage}. To avoid evaluating on the same set of images used for extracting the explanations, we use the 50k-image Imagenet test set, $\mathcal{D}^{\text{test}}$. Finally, building on top of recent work~\citep{shaham2024multimodal,bai2024describeanddissect}, we propose a pair of metrics based on synthetic images generated by diffusion model.

\paragraph{Simulation-based Evaluation.}
\citet{zhang2024largemultimodalmodelsinterpret,xu2025decipheringfunctionsneuronsvisionlanguage} propose using a segmentation model $m_{\text{seg}}$, (e.g., SAM 2~\citep{ravi2025sam}) to generate binary masks $M_{\text{seg}}$ containing 1s on the image patches that correspond to the concepts described in the explanations. These masks simulate how the SAE feature would activate if the explanation were true. $M_{\text{seg}}$ masks are compared against the actual feature's activation masks $M_{\text{feature}}$. More formally, given an image and an explanation, the masks are computed as follows:
\begin{equation}
M^{i,I}_{\text{feature}} = \mathds{1}[f_i(m_{\text{sub}}(I)) > 0], \quad M^{i,I}_{\text{seg}} = m_{\text{seg}}(I, \mathbf{e}_i),
\end{equation}
where $\mathds{1}[\cdot]$ is an indicator function that returns 1 if the condition holds and 0 otherwise. To quantitatively assess the alignment between these simulated and actual activation masks, the Intersection over Union (IoU) is computed and averaged over the top-k activating images $\mathcal{T}^k_i$ on $\mathcal{D}^{\text{test}}$:
\begin{equation}\label{eq:iou_score}
\text{IoU-Score}^i = \frac{1}{k} \sum_{I \in \mathcal{T}^k_i} 
 \frac{|M^{i,I}_{\text{seg}} \cap M^{i,I}_{\text{feature}}|}{|M^{i,I}_{\text{seg}} \cup M^{i,I}_{\text{feature}}|}.
\end{equation}

\paragraph{CLIP-based Evaluation.}
To assess the semantic alignment between explanations and the corresponding top-$k$ activating images in $\mathcal{D}^{\text{test}}$, we follow~\citet{zhang2024largemultimodalmodelsinterpret} and use a CLIP model $m_{\text{clip}}$. For each dimension $i$, we compute the text embedding from the explanation $\mathbf{e}_i$ and extract visual embeddings from the top-$k$ activating images $\mathcal{T}^k_i$ associated with that feature. Specifically, for each image $I \in \mathcal{T}^k_i$, we apply the feature's activation masks ($M^{i,I}_{\text{feature}}$) to focus on the relevant region, and compute its CLIP image embedding. We then measure the cosine similarity between the explanation embedding and each masked image embedding, averaged across images:
\begin{equation}
\text{CLIP-Score}^i = \frac{1}{k} \sum_{I \in \mathcal{T}^k_i} \cos\left(m_{\text{clip}}^{\text{text}}(\mathbf{e}_i), m_{\text{clip}}^{\text{img}}(I)\right).
\end{equation}

\begin{table*}[t]
\caption{Explanation evaluation metrics for the middle layer SAE of Gemma 3 and InternVL3-14B vision encoders. Except for AUROC, mean scores are reported, and statistical significance is assessed pairwise between methods. A value is underlined if it is significantly higher (with $p<0.05$) than both other methods in the same column.}
\centering
\renewcommand{\arraystretch}{1.3}  
\resizebox{0.87\textwidth}{!}{%
\begin{tabular}{c c *{8}{c}}
\toprule
\multirow{2}{*}{\textbf{Model}}
& \multirow{2}{*}{\textbf{Explanation Method}} 
& \multicolumn{2}{c}{\textbf{IoU Score}} 
& \multicolumn{2}{c}{\textbf{AUROC}} 
& \multicolumn{2}{c}{\textbf{Synth. Act. Score}} 
& \multicolumn{2}{c}{\textbf{CLIP Score}} \\
\cmidrule(lr){3-4} \cmidrule(lr){5-6} \cmidrule(lr){7-8} \cmidrule(lr){9-10}
& & Masks & Heatmaps & Masks & Heatmaps & Masks & Heatmaps & Masks & Heatmaps \\
\midrule
\multirow{3}{*}{\raisebox{-1.5em}{\rotatebox{90}{Gemma 3}}}
& \textbf{Steering} 
& \multicolumn{2}{c}{0.211} 
& \multicolumn{2}{c}{0.675} 
& \multicolumn{2}{c}{0.324} 
& \multicolumn{2}{c}{0.186} \\
& \textbf{Top-k}    
& 0.211 & 0.198 
& 0.723 & 0.791 
& 0.330 & 0.364 
& 0.190 & 0.187 \\
& \textbf{Steering-informed Top-k} 
& 0.216 & 0.203
& 0.788 & 0.838
& \underline{0.461} & \underline{0.505}
& \underline{0.193} & \underline{0.189} \\
\midrule
\multirow{3}{*}{\raisebox{-1.5em}{\rotatebox{90}{InternVL3}}}
& \textbf{Steering} 
& \multicolumn{2}{c}{0.220} 
& \multicolumn{2}{c}{0.655} 
& \multicolumn{2}{c}{0.141} 
& \multicolumn{2}{c}{0.191} \\
& \textbf{Top-k}    
& 0.224 & 0.201 
& 0.768 & 0.775 
& 0.187 & 0.183 
& 0.199 & 0.187 \\
& \textbf{Steering-informed Top-k} 
& 0.228 & 0.203 
& 0.823 & 0.833
& \underline{0.254} & \underline{0.252} 
& \underline{0.199} & 0.191 \\
\bottomrule
\end{tabular}%
}
\label{tab:merged_mid_masks_heatmaps}
\end{table*}

\paragraph{Synthetic-image-based Evaluation.}
For each feature $i$, we  generate a set of $N$ \textit{positive} images using a diffusion model\footnote{We use Stable Diffusion 3.5 Medium~\citep{esser2024scalingrectifiedflowtransformers}.} $m_{\text{diff}}$ conditioned on the explanation $\mathbf{e}_i$, ${\mathcal{I}^{i, +} = \left\{I \sim m_{\text{diff}}(I \mid \mathbf{e}_i) \right\}^{N}}$. Then, we compute the average feature (synthetic)~image activation score~(\Cref{eq:activation_score_image}):
\begin{equation}\label{eq:synth_act_score}
    \text{Synthetic-Activation-Score}^i = \frac{1}{N} \sum_{I \in \mathcal{I}^{i,+}} S^{I,i}.
\end{equation}
We also generate a set of $N$ \textit{negative} images, $\mathcal{I}^{i, -} = \left\{ I \sim \mathcal{D}^{\text{test}} \right\}^{N}$ by randomly sampling from the test set. Following~\Cref{eq:activation_score_image}, we obtain the \textit{image activation score} for each positive and negative image and repeat the process for every feature. Finally, we compute the AUROC metric.\footnote{This is mathematically equivalent to the probability that the obtained image activation score for a `positive' image in $\mathcal{I}^{i, +}$--generated by $m_{\text{diff}}$--ranks higher than a randomly chosen negative image from $\mathcal{D}^{\text{test}}$.}

\subsection{Experimental Setup}
We train SAEs on a middle-layer of the vision encoders of Gemma 3~\citep{gemmateam2024gemma2improvingopen} and the InternVL3-14B~\citep{zhu2025internvl3}, two state-of-the-art VLMs. We also train a SAE at a later layer~(3/4th depth) of Gemma 3 encoder. Gemma 3 employs a 400M parameters variant of the SigLIP encoder~\citep{zhai2023sigmoidlosslanguageimage}, which works at a fixed resolution of \(896 \times 896\) pixels. It remains frozen during LM training and adaptation stages and produces 4,096 tokens per image. In contrast, InternVL3-14B incorporates the pretrained InternViT-300M-448px-V2\_5 encoder (300M parameters), which processes images at a \(448 \times 448\) resolution, producing 256 tokens per input. This setup enables us to evaluate our proposed methods on a `pure' SigLIP encoder~(Gemma 3) and another encoder adapted through joint training (InternVL3).

Unless stated otherwise, the explainer models correspond to the same VLM from which the encoder is interpreted, Gemma 3 27B and InternVL3-14B respectively. The prompts used for \textit{Top-$k$} and \textit{Steering-informed Top-$k$} (these two methods share the same prompt) are designed to closely mirror that of \steering, ensuring consistency across methods (see~\Cref{apx:prompts}). For all experiments involving top-activating images, we report results using the top five images~(i.e., $k$=5). 
\section{Results}
\begin{figure*}[!t]
\centering
\begin{minipage}{.53\textwidth}
  \centering
  \includegraphics[width=0.91\textwidth]{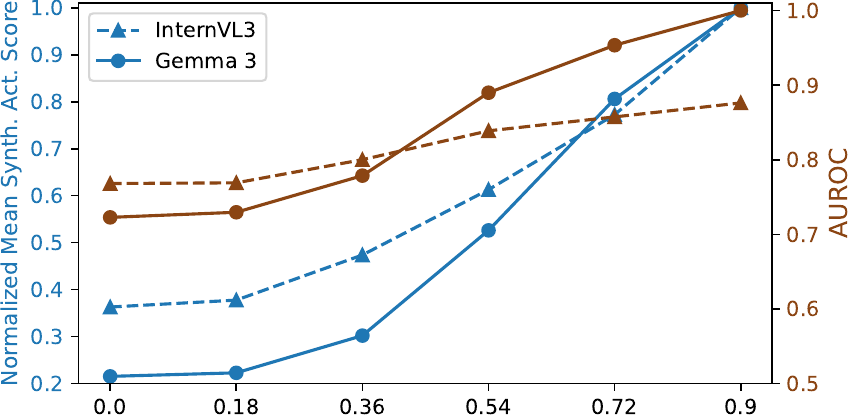}
    \caption{Middle layer SAE synthetic-image-based evaluation scores of \topk method  as a function of the similarity with \steering Explanations.}
\label{fig:gemma_vs_intern_sim_threshold_gen_img_mean_mean_gen_img_mean_auroc}
\end{minipage}%
\hfill
\begin{minipage}{.44\textwidth}
  \centering
  \includegraphics[width=.85\linewidth]{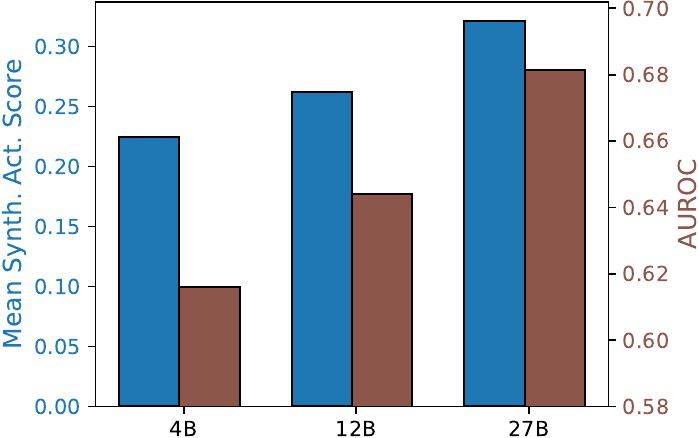}
  \caption{Gemma 3 synthetic-image-based evaluation scores of \steering method as a function of the size of the LM $\text{m}_{\text{subj}}$.}
  \label{fig:activation_auroc_plot_dataset_size_mean_decoder_size}
\end{minipage}
\end{figure*}

To compare the different explanation methods we evaluate the quality of the explanations generated by these methods using the metrics described in~\Cref{sec:evaluation_methods}.
Our analysis is divided in three parts. Section~\ref{sec:steering_results} evaluates the performance of the \steering method and illustrates its potential to reduce \textit{contextual bias} present in standard \topk explanations. Section~\ref{sec:improving_topk} shifts focus to the \stinformed method, showing how it improves explanation quality. Finally, Section~\ref{sec:feature_space} explores the SAE feature space to uncover the semantic structure of learned features.



\subsection{Explaining through Steering} \label{sec:steering_results} 

We analyze the effectiveness of the \steering method focusing on how it scales with model size, performs across different evaluation metrics, and complements \topk explanations. Our results suggest that, although \steering has limitations when used in isolation, it scales effectively, inherently mitigates \textit{contextual biases}, and can be used to improve other interpretability methods, despite being more efficient.

\paragraph{Steering performs well on IoU but lags behind in the rest of metrics.} While \steering performs competitively on IoU Score, especially on later layer SAE~(\Cref{apx:later_layer}), it consistently lags behind in the remaining metrics. This is evident across all evaluated layers and models, where \steering achieves solid overlap with segmentation masks but fails to elicit strong activations or achieve high AUROC and CLIP alignment. In particular, its AUROC and Mean Synthetic Activation scores are substantially lower than those of \stinformed, indicating weaker model sensitivity and less effective explanation quality. To verify that steering performance arises from feature-specific explanations rather than driven by the prompt or explained by randomness, we compare against random norm-matched vectors and permuted-direction controls, where randomly sampled SAE features are selected instead. As shown in~\Cref{apx:faithfulness_control}, these baselines perform substantially worse across all metrics, supporting the causal relevance of the injected feature directions.

\paragraph{Steering helps surface high-quality Top-k explanations.} We hypothesize that the \steering method can act as a valuable signal for validating \topk explanations. Intuitively, if both methods independently produce semantically similar explanations for the same feature, this agreement may indicate a higher likelihood of correctness. To test this, we compute semantic embeddings of each explanation using a sentence similarity model~\citep{reimers-gurevych-2019-sentence} and measure the semantic similarity between the explanations produced by \topk and \steering. We then assess the quality of the \topk explanations as a function of this similarity, retaining only those above varying thresholds.

As shown in \Cref{fig:gemma_vs_intern_sim_threshold_gen_img_mean_mean_gen_img_mean_auroc}, explanation quality—measured by normalized synthetic activation scores and AUROC—improves consistently as the similarity to \steering increases. This trend holds across both the Gemma 3 and InternVL3 encoders with the exception of Gemma's CLIP Score (Appendix~\ref{apx:iou_clip_sim}). These results suggest that \steering serves as an effective filter or guide, helping to identify high-quality explanations and improving the overall interpretability pipeline when used in conjunction with \topk.

\paragraph{Steering quality scales with LM size. } \steering explanations improve as the size of the underlying language model used for generation increases. In this experiment, we vary the size of the LM used to produce explanations while keeping all other components fixed. As shown in \Cref{fig:activation_auroc_plot_dataset_size_mean_decoder_size}, both evaluation metrics—Mean Synthetic Activation Score and AUROC—show consistent improvements when moving from 4B to 12B to 27B parameter models. A positive trend is also observed for the rest of the metrics in~\Cref{apx:iou_clip_function_dec_size}. This suggests that larger language models generate more informative and causally effective explanations when used in the \steering framework. Crucially, this trend points to a promising direction: as language models continue to grow in scale and capability, we can expect the quality of \steering-based interpretability to improve accordingly.

\begin{figure*}[!t]
\vspace{-15pt}
\centering
\begin{minipage}[t]{0.48\textwidth}
\centering
\vspace{15pt}
\setlength{\tabcolsep}{3pt} 
\renewcommand{\arraystretch}{1.525} 
\resizebox{\textwidth}{!}{%
\begin{tabular}{c l c c c c c c}
\toprule
\textbf{Model}
& \textbf{Masking Type}
& \multicolumn{2}{c}{\shortstack{\textbf{Steering} \\ ~}} 
& \multicolumn{2}{c}{\shortstack{\textbf{Top-k} \\ ~}} 
& \multicolumn{2}{c}{\shortstack{\textbf{Steering-informed} \\ \textbf{Top-k}}} \\
\cmidrule(lr){3-4} \cmidrule(lr){5-6} \cmidrule(lr){7-8}
& & \textbf{Count} & \textbf{\%} & \textbf{Count} & \textbf{\%} & \textbf{Count} & \textbf{\%} \\
\midrule
\multirow{2}{*}{\rotatebox[origin=c]{90}{\scriptsize Gemma 3}} 
& Masks    & 0 & 0.0\% & 23  & 7.7\%  & 12  & 4.0\% \\
& Heatmaps & 0 & 0.0\% & 125 & 47.7\% & 116 & 38.6\% \\
\midrule
\multirow{2}{*}{\rotatebox[origin=c]{90}{\scriptsize InternVL3}} 
& Masks    & 0 & 0.0\% & 19  & 6.3\%  & 13  & 4.3\% \\
& Heatmaps & 0 & 0.0\% & 125 & 41.7\% & 117 & 39.0\% \\
\bottomrule
\end{tabular}
}
\vspace{3pt}
\captionof{table}{Count and percentage of `background' explanations turned `animal' explanations by different methods~(see main text for details).}
\label{tab:directional_transitions}
\end{minipage}
\hfill
\begin{minipage}[t]{0.5\textwidth}
\vspace{10pt}
\centering
\includegraphics[width=0.8\textwidth]{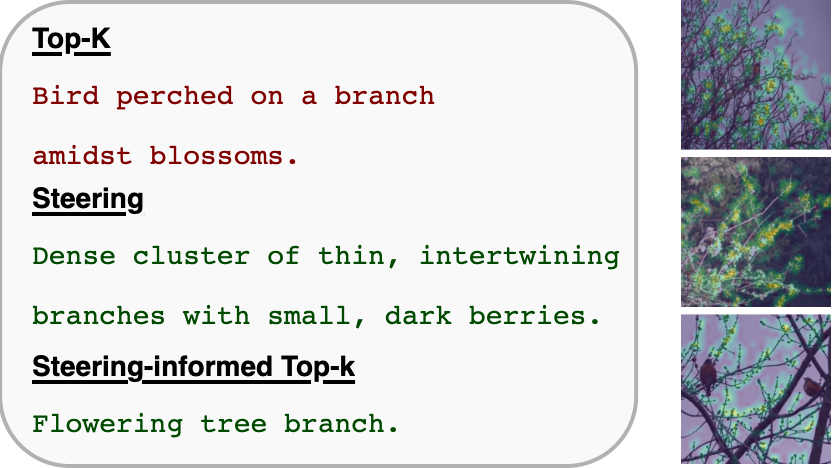}
\caption{Example of \topk explanation exhibiting \textit{contextual bias}.}
\label{fig:contextual_bias_example}
\end{minipage}
\end{figure*}

\begin{figure*}[!t]
\begin{centering}
\begin{minipage}[t]{0.46\textwidth}
    \centering
    \includegraphics[width=\linewidth]{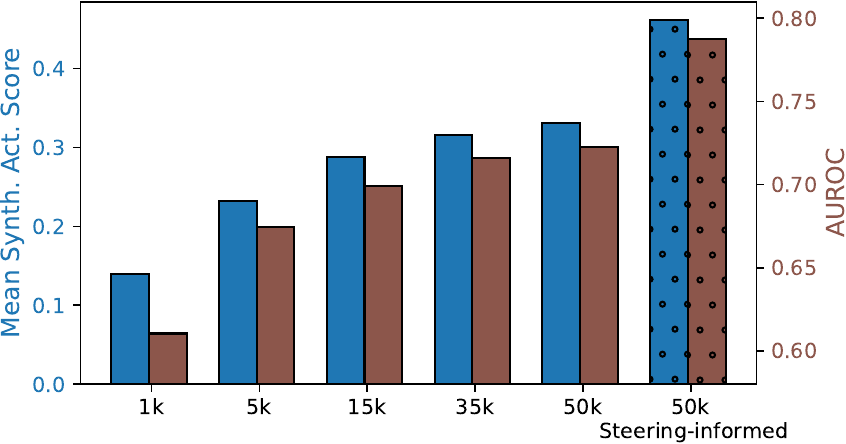}
    \label{fig:mean_auroc_dataset_size}
\end{minipage}
\hfill
\begin{minipage}[t]{0.469\textwidth}
    \centering
    \includegraphics[width=\linewidth]{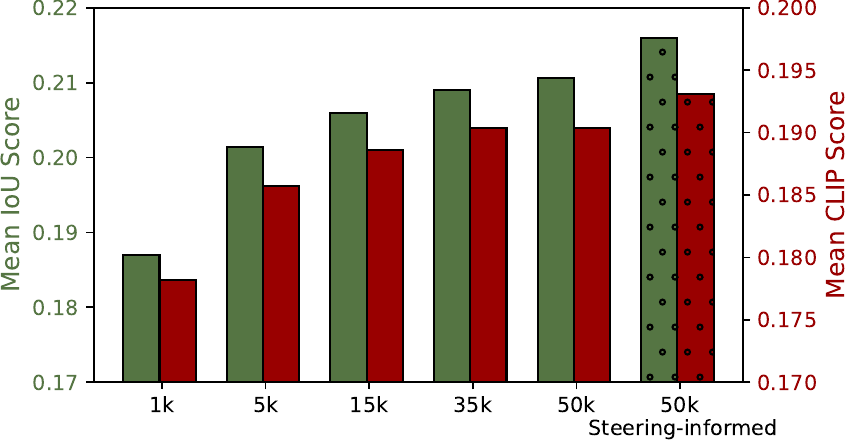}
    \label{fig:iou_clip_plot_dataset_size_mean}
\end{minipage}
\vspace{-10pt}
\caption{Explanation evaluation scores--synthetic-image-based scores on the left, and IoU and CLIP scores on the right of \topk method as a function of the evaluation set size. \stinformed results on the rightmost bar.}
\end{centering}
\end{figure*}

\paragraph{Steering prevents contextual biases found in Top-k.} To better understand what \steering captures that \topk does not, we analyze the 300 features with the largest IoU score difference between the two methods. Manual inspection of this subset reveals that \steering often produces accurate \textit{background} explanations, whereas \topk tends to misattribute these features to foreground elements such as animals, likely due to recurring context in the top activating images, a pattern we name \textit{contextual bias}~(see~\Cref{fig:figure_1,fig:contextual_bias_example}).

To quantify this effect, we categorize each explanation using Gemma 3 27B as \textit{background}, \textit{animal}, or \textit{other}. As shown in~\Cref{tab:directional_transitions}, \topk explanations frequently fall into the \textit{animal} category (e.g., 47.7\% with heatmaps), despite the feature aligning with \textit{background} under \steering with a high IoU score. Notably, the hybrid \stinformed reduces this misattribution (to 38.6\%), suggesting it inherits some of \steering’s robustness to contextual bias.

\subsection{The Best of Both Worlds: \stinformed} \label{sec:improving_topk}

We now analyze how the \stinformed method consistently improves explanation quality. In this section, we highlight two key findings: the consistent superiority of \stinformed across all metrics, and its ability to overcome the diminishing returns when using larger datasets.

\paragraph{Steering-informed Top-k gives the best explanations across the board.} Across models and layers, \stinformed consistently achieves the best performance across all evaluation metrics—IoU Score, AUROC, Mean Synthetic Activation, and CLIP Score—demonstrating its superiority in producing high-quality explanations. In both the middle and later layers of the Gemma 3 vision encoder~(\Cref{tab:merged_mid_masks_heatmaps} top, and \Cref{tab:gemma3_later_masks_heatmaps}), as well as in the middle layer of the InternVL3-14B encoder (\Cref{tab:merged_mid_masks_heatmaps} bottom), this method outperforms both standard \steering and \topk approaches, regardless of whether masks or heatmaps are used. Notably, it achieves the highest AUROC and Synthetic Activation scores, indicating that the explanations not only align well with segmentation and top-$k$ activating images, but also elicit stronger feature activations when using synthetic examples. These results underline the effectiveness of combining top-k selection with causal interventions to enhance explanation quality.

\begin{figure*}[!t]
\begin{centering}
    \includegraphics[width=0.95\textwidth]{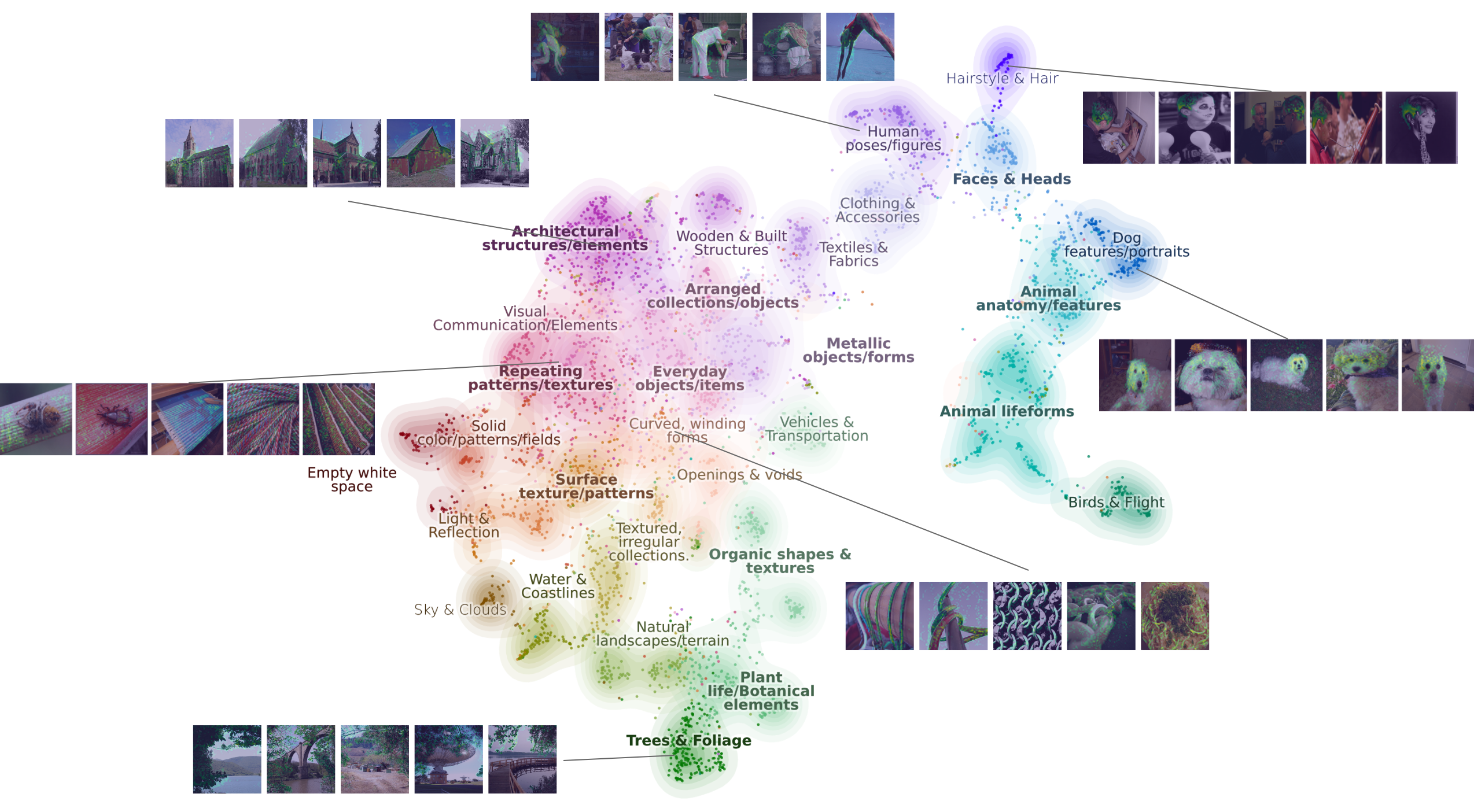}
    \caption{UMAP visualization of SAE feature explanations, Gemma 3 vision encoder, middle layer.}
    \label{fig:datamapplot_google_gemma-3-27b-it_mid_masks}
\end{centering}
\end{figure*}

\paragraph{Steering-informed Top-k overcomes diminishing returns.} We additionally generate \topk and \stinformed explanations using the top-k images obtained with a reduced evaluation dataset. As observed in~\Cref{fig:mean_auroc_dataset_size,fig:iou_clip_plot_dataset_size_mean}, as the size of the evaluation dataset increases, standard \topk explanations gradually improve in quality, but the gains exhibit diminishing returns, especially beyond 15k examples. This trend is visible across all metrics. In contrast, \stinformed provides an immediate and substantial performance boost, effectively bypassing the need for large-scale data to reach high-quality explanations, with particular improvements in synthetically generated metrics~(\Cref{fig:mean_auroc_dataset_size}), suggesting that the causal intervention adds valuable signal beyond what dataset scaling alone can offer.

\subsection{Exploring the SAE Feature Space}\label{sec:feature_space}

To complement the previous evaluations, this section provides an overview of the structure of the learned SAE feature space. For this purpose, we use the middle-layer SAE of Gemma 3 encoder.

\paragraph{Selecting the best explanation per feature.} Inspired by~\citet{choi2024automatic}, which identifies the best explanations from a set of candidates, we adopt a rank-based voting strategy to select the top explanation across the three explanation methods for each SAE feature. Specifically, each evaluation metric independently ranks each explanation method. Then, the explanation with the lowest (best) total rank is selected. In case of a tie, the explanation is chosen at random.

\begin{figure*}[!t]
\centering
\begin{minipage}[t]{0.48\textwidth}
\begin{centering}
    \includegraphics[width=\textwidth]{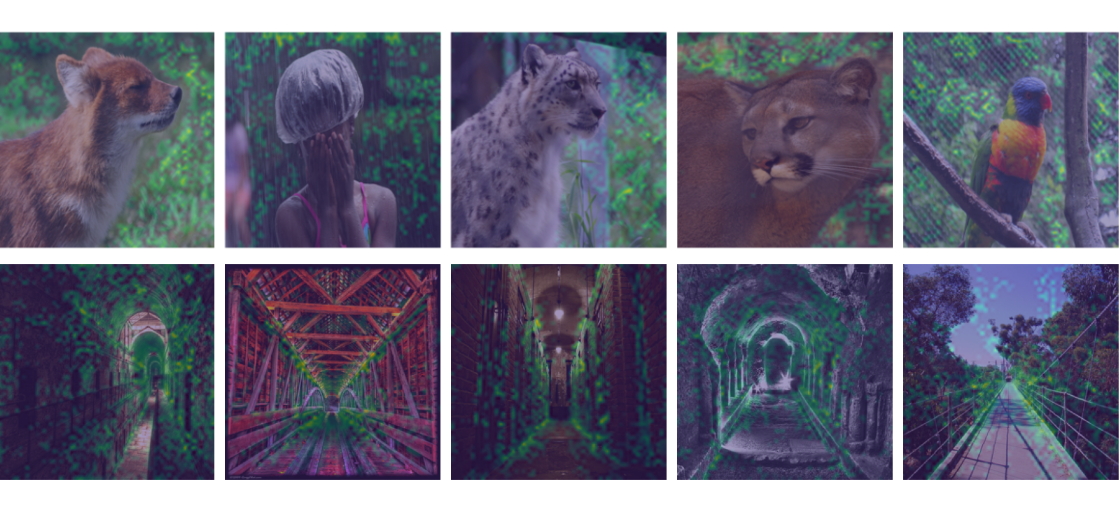}
   \caption{\textit{Depth}~(top) and \textit{perspective}~(bottom) features previously found as unique to Dinov2, surfaced via steering explanations.}
    \label{fig:joint_no_only_dino_features}
\end{centering}
\end{minipage}
\hfill
\begin{minipage}[t]{0.47\textwidth}
\centering
\includegraphics[width=\textwidth]{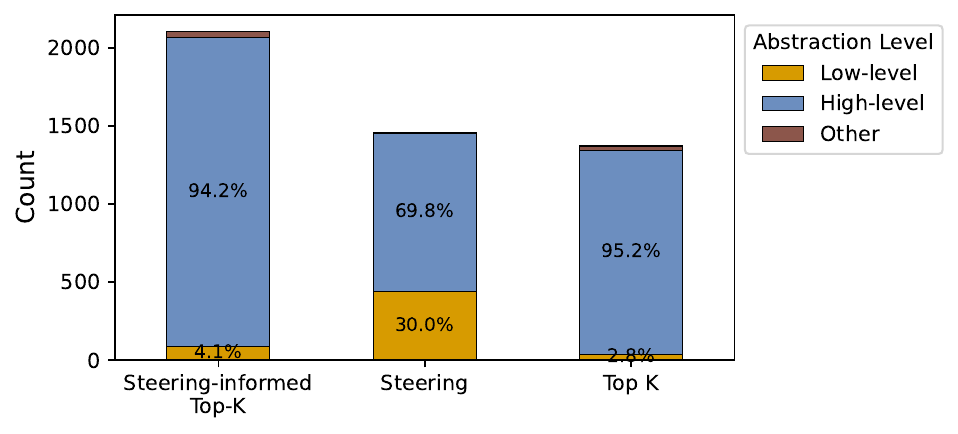}
\caption{Count of selected explanations by each method. Each bar shows the level of abstraction of the selected explanations.}
\label{fig:explanation_distribution_abstraction}
\end{minipage}
\end{figure*}

To ensure we select interpretable features with meaningful explanations, we discard features whose selected explanation has an IoU score~(\Cref{eq:iou_score}) below 0.2 or synthetic activation score~(\Cref{eq:synth_act_score}) below 0.3. This filtering step leaves around 5,000 out of the original 7,690 alive features with assigned explanation. As shown in~\Cref{fig:explanation_distribution_abstraction}, \stinformed is selected more frequently. Notably, \steering and \topk explanations are selected at similar rates.

\paragraph{Visualizing the SAE feature space.} After selecting high-quality explanations, we compute semantic embeddings using a sentence similarity model\footnote{We use \href{https://huggingface.co/sentence-transformers/all-mpnet-base-v2}{\texttt{sentence-transformers/all-mpnet-base-v2}}
.}~\citep{reimers-gurevych-2019-sentence}. The projected 2D UMAP~\citep{mcinnes2018umap-software} representation of these embeddings is shown in~\Cref{fig:datamapplot_google_gemma-3-27b-it_mid_masks}, where the clusters are obtained via k-means algorithm~\citep{lloyd1982least} with $k=30$. To facilitate interpretation, we assign a label to each cluster by giving Gemma 3 27B a random sample of 20 explanations from that cluster.

Since the SAE is trained on ImageNet, the learned features seem to capture concepts prevalent in the dataset, such as humans~(\textit{Human poses/figures}), animals~(\textit{Animal lifeforms}), and natural scenes (\textit{Trees \& Foliage}). While many explanations correspond to high-level semantic categories (\textit{e.g., } \textit{Vehicles}, \textit{Clothing}), which aligns with expectations for middle-layer features~\citep{cammarata2020thread}, we also observe features at lower levels of abstraction. These include perceptual features like \textit{Repeating patterns/textures} and  \textit{Surface texture/patterns}. Notably, as shown in Figure \ref{fig:explanation_distribution_abstraction}, \steering allows obtaining these low-level features.

\paragraph{Finding features previously thought unique to DinoV2.} The semantic space of explanation embeddings enables targeted retrieval of features aligned with user-specified concepts. As a proof of concept, we search for features previously identified by~\citet{thasarathan2025universalsparseautoencodersinterpretable} as unique to DinoV2~\citep{oquab2024dinov}, a vision model trained without language supervision. Contrary to prior claims, we found features seemingly representing \textit{depth}~(\Cref{fig:joint_no_only_dino_features} top) and \textit{perspective}~(\Cref{fig:joint_no_only_dino_features} bottom) in our SigLIP SAE. For instance, the \textit{depth} feature is described by the \steering explanation as: “\textit{Blurred, out-of-focus background creating a sense of depth and indistinctness.}”, and the \textit{perspective} feature as “\textit{Long, receding perspective created by converging lines, evoking a sense of depth and distance}”. While anecdotal, these findings demonstrate the utility of combining steering-based explanations with semantic search to uncover conceptual overlap across models.
\section{Related Work}


Interpretability in vision models has seen rapid progress~\citep{fel2026rabbithulltaskrelevantconcepts,Joseph2024,joseph2025prismaopensourcetoolkit}, with recent work aiming at mapping internal representations to natural language. A key strategy has been to leverage CLIP's shared image-text embedding space to align vision model features with human-understandable concepts~\citep{gandelsman2023interpreting,bhalla2024interpretingclipsparselinear}.

In parallel, mechanistic interpretability has advanced our understanding of LLMs~\citep{ferrando2024primerinnerworkingstransformerbased}, with SAEs revealing interpretable features~\citep{bricken2023monosemanticity}. Recently, SAEs have been applied to vision models~\citep{sae_vision,lim2025sparse,thasarathan2025universalsparseautoencodersinterpretable,rajaram2025linesightlinearrepresentations,venhoff2025visualrepresentationsmaplanguage,shabalin2025interpretinglargetexttoimagediffusion,gorton2024missingcurvedetectorsinceptionv1,surkov2025onestep}, revealing semantically meaningful features. Yet, interpreting thousands of features remains a bottleneck, highlighting the need for automated solutions.

Automated interpretability in LLMs has traditionally followed `input-centric' strategies, where explanations are generated from top-activating inputs~\citep{openai_neuron_nle, choi2024automatic}. This input-centric method perspective has been extended to vision SAEs~\citep{zhang2024largemultimodalmodelsinterpret,xu2025decipheringfunctionsneuronsvisionlanguage,10.1007/978-3-031-72980-5_26}, where top-activating images are used instead. To address input-centric limitations, recent work has shifted toward output-centric explanations.~\citet{gurarieh2025enhancingautomatedinterpretabilityoutputcentric} propose \textit{VocabProj} and \textit{TokenChange} to reveal which outputs are causally tied to specific features. Similarly, \citet{paulo2024automaticallyinterpretingmillionsfeatures} introduce an intervention-based metric to assess explanation quality through causal influence. In vision models, output-centric causal approaches based on steering have also emerged, though applications have so far remained confined to within-model interventions~\citep{joseph2025steeringclipsvisiontransformer, lim2025sparse, stevens2025sparseautoencodersscientificallyrigorous}, while we propose leveraging a language model to generate the explanation on the intervened vision encoder.

Closely related to our work are efforts on self-explaining features in LLMs. \textit{Patchscopes}~\citep{ghandeharioun2024patchscopes,chen2024selfie} use activation patching to transfer representations and generate causal explanations. \citet{self_explaining_saes} extend this idea to SAEs, enabling the model to act as its own explainer by describing its features. More recently, this line of work has shifted toward \textit{training} models to better verbalize their internal representations~\citep{pan2026latentqa,li2026traininglanguagemodelsexplain,karvonen2026activationoraclestrainingevaluating}. These approaches enable more robust interpretation of model features, but require additional supervision. Extending such training-based approaches to vision-language models represents a promising direction for future work.

\section{Conclusions}

This work presents a new framework for automatically interpreting features in vision models. By steering the encoder with targeted feature interventions alone, and leveraging a language model as the explainer, we generate feature explanations in an efficient and scalable way. While \steering overall tends to underperform \topk method, it avoids their contextual biases and is particularly effective at surfacing lower-level features. Moreover, combining both approaches enables the identification of higher-quality explanations, highlighting their complementary nature. Explanation quality also scales consistently with language model size, suggesting that as LMs continue to advance, steering-based explanations will become increasingly informative and precise. The hybrid \stinformed~approach consistently produces the highest-quality explanations across evaluation metrics, demonstrating the value of integrating causal interventions with input-based methods.

\section{Acknowledgments}
During the development of this work, Javier Ferrando, Pablo Agustin Martin-Torres, Daniel Hinjos and Anna Arias Duart were supported by the fellowship within the ``Generación D'' initiative, \href{https://www.red.es/es}{Red.es}, Ministerio para la Transformación Digital y de la Función Pública, for talent attraction (C005/24-ED CV1). Funded by the European Union NextGenerationEU funds, through PRTR. This work was also supported by the ELLIOT Grant, funded by the European Union under grant agreement No. 10121439.

{
    \small
    \bibliographystyle{ieeenat_fullname}
    \bibliography{main}

@String(ECCV= {Eur. Conf. Comput. Vis.})

@String(ICLR = {Int. Conf. Learn. Represent.})

@String(ECCV  = {ECCV})

@String(ICLR  = {ICLR})

@article{elhage2022superposition,
    title={Toy Models of Superposition},
    author={Elhage, Nelson and Hume, Tristan and Olsson, Catherine and Schiefer, Nicholas and Henighan, Tom and Kravec, Shauna and Hatfield-Dodds, Zac and Lasenby, Robert and Drain, Dawn and Chen, Carol and Grosse, Roger and McCandlish, Sam and Kaplan, Jared and Amodei, Dario and Wattenberg, Martin and Olah, Christopher},
    year={2022},
    journal={Transformer Circuits Thread},
    url={https://transformer-circuits.pub/2022/toy_model/index.html}
}

@article{bricken2023monosemanticity,
    title={Towards Monosemanticity: Decomposing Language Models With Dictionary Learning},
    author={Bricken, Trenton and Templeton, Adly and Batson, Joshua and Chen, Brian and Jermyn, Adam and Conerly, Tom and Turner, Nick and Anil, Cem and Denison, Carson and Askell, Amanda and Lasenby, Robert and Wu, Yifan and Kravec, Shauna and Schiefer, Nicholas and Maxwell, Tim and Joseph, Nicholas and Hatfield-Dodds, Zac and Tamkin, Alex and Nguyen, Karina and McLean, Brayden and Burke, Josiah E and Hume, Tristan and Carter, Shan and Henighan, Tom and Olah, Christopher},
    year={2023},
    journal={Transformer Circuits Thread},
    url={https://transformer-circuits.pub/2023/monosemantic-features/index.html}
}

@book{pearl_2009,
    place={Cambridge},
    edition={2},
    title={Causality},
    doi={10.1017/CBO9780511803161},
    publisher={Cambridge University Press},
    author={Pearl, Judea},
    year={2009}
}

@article{olah2020zoom,
    author = {Olah, Chris and Cammarata, Nick and Schubert, Ludwig and Goh, Gabriel and Petrov, Michael and Carter, Shan},
    title = {Zoom In: An Introduction to Circuits},
    journal = {Distill},
    year = {2020},
    url = {https://distill.pub/2020/circuits/zoom-in},
    doi = {10.23915/distill.00024.001}
}

@article{cammarata2020thread,
    author = {Cammarata, Nick and Carter, Shan and Goh, Gabriel and Olah, Chris and Petrov, Michael and Schubert, Ludwig and Voss, Chelsea and Egan, Ben and Lim, Swee Kiat},
    title = {Thread: Circuits},
    journal = {Distill},
    year = {2020},
    url = {https://distill.pub/2020/circuits},
    doi = {10.23915/distill.00024}
}

@article{ghandeharioun2024patchscopes,
    title={Patchscopes: A Unifying Framework for Inspecting Hidden Representations of Language Models}, 
    author={Asma Ghandeharioun and Avi Caciularu and Adam Pearce and Lucas Dixon and Mor Geva},
    year={2024},
    journal={Arxiv},
    url={https://arxiv.org/abs/2401.06102v2}
}

@misc{openai_neuron_nle,
  title = {Language models can explain neurons in language models},
  author = {Steven Bills and Nick Cammarata and Dan Mossing and Henk Tillman and Leo Gao and Gabriel Goh and Ilya Sutskever and Jan Leike and Jeff Wu and William Saunders},
  year = {2023},
  howpublished = {\url{https://openaipublic.blob.core.windows.net/neuron-explainer/paper/index.html}},
}

@misc{kaplan2020scaling,
      title={Scaling Laws for Neural Language Models}, 
      author={Jared Kaplan and Sam McCandlish and Tom Henighan and Tom B. Brown and Benjamin Chess and Rewon Child and Scott Gray and Alec Radford and Jeffrey Wu and Dario Amodei},
      year={2020},
      eprint={2001.08361},
      archivePrefix={arXiv},
      primaryClass={cs.LG},
        url={https://arxiv.org/abs/2001.08361}
}

@article{sae_vision,
    author = {Hugo Fry},
    title = {Towards Multimodal Interpretability: Learning Sparse Interpretable Features in Vision Transformers},
    journal={LessWrong},
    URL = {https://www.lesswrong.com/posts/bCtbuWraqYTDtuARg/towards-multimodal-interpretability-learning-sparse-2},
    year = {2024}
}

@misc{chen2024selfie,
      title={SelfIE: Self-Interpretation of Large Language Model Embeddings}, 
      author={Haozhe Chen and Carl Vondrick and Chengzhi Mao},
      year={2024},
      eprint={2403.10949},
      archivePrefix={arXiv},
      primaryClass={cs.CL},
        url={https://arxiv.org/abs/2403.10949}
}

@article{shaham2024multimodal,
      title={A Multimodal Automated Interpretability Agent}, 
      author={Tamar Rott Shaham and Sarah Schwettmann and Franklin Wang and Achyuta Rajaram and Evan Hernandez and Jacob Andreas and Antonio Torralba},
      journal={Arxiv},
      year={2024},
      url={https://arxiv.org/abs/2404.14394}
}

@article{zhu2025internvl3,
  title={InternVL3: Exploring Advanced Training and Test-Time Recipes for Open-Source Multimodal Models},
  author={Zhu, Jinguo and Wang, Weiyun and Chen, Zhe and Liu, Zhaoyang and Ye, Shenglong and Gu, Lixin and Duan, Yuchen and Tian, Hao and Su, Weijie and Shao, Jie and others},
  journal={arXiv preprint arXiv:2504.10479},
  year={2025}
}

@inproceedings{
gao2025scaling,
title={Scaling and evaluating sparse autoencoders},
author={Leo Gao and Tom Dupre la Tour and Henk Tillman and Gabriel Goh and Rajan Troll and Alec Radford and Ilya Sutskever and Jan Leike and Jeffrey Wu},
booktitle={The Thirteenth International Conference on Learning Representations},
year={2025},
url={https://openreview.net/forum?id=tcsZt9ZNKD}
}

@article{gemmateam2024gemma2improvingopen,
      title={Gemma 2: Improving Open Language Models at a Practical Size}, 
      author={Gemma Team and Morgane Riviere and Shreya Pathak and Pier Giuseppe Sessa and Cassidy Hardin and Surya Bhupatiraju and Léonard Hussenot and Thomas Mesnard and Bobak Shahriari and Alexandre Ramé and Johan Ferret and Peter Liu and Pouya Tafti and Abe Friesen and Michelle Casbon and Sabela Ramos and Ravin Kumar and Charline Le Lan and Sammy Jerome and Anton Tsitsulin and Nino Vieillard and Piotr Stanczyk and Sertan Girgin and Nikola Momchev and Matt Hoffman and Shantanu Thakoor and Jean-Bastien Grill and Behnam Neyshabur and Olivier Bachem and Alanna Walton and Aliaksei Severyn and Alicia Parrish and Aliya Ahmad and Allen Hutchison and Alvin Abdagic and Amanda Carl and Amy Shen and Andy Brock and Andy Coenen and Anthony Laforge and Antonia Paterson and Ben Bastian and Bilal Piot and Bo Wu and Brandon Royal and Charlie Chen and Chintu Kumar and Chris Perry and Chris Welty and Christopher A. Choquette-Choo and Danila Sinopalnikov and David Weinberger and Dimple Vijaykumar and Dominika Rogozińska and Dustin Herbison and Elisa Bandy and Emma Wang and Eric Noland and Erica Moreira and Evan Senter and Evgenii Eltyshev and Francesco Visin and Gabriel Rasskin and Gary Wei and Glenn Cameron and Gus Martins and Hadi Hashemi and Hanna Klimczak-Plucińska and Harleen Batra and Harsh Dhand and Ivan Nardini and Jacinda Mein and Jack Zhou and James Svensson and Jeff Stanway and Jetha Chan and Jin Peng Zhou and Joana Carrasqueira and Joana Iljazi and Jocelyn Becker and Joe Fernandez and Joost van Amersfoort and Josh Gordon and Josh Lipschultz and Josh Newlan and Ju-yeong Ji and Kareem Mohamed and Kartikeya Badola and Kat Black and Katie Millican and Keelin McDonell and Kelvin Nguyen and Kiranbir Sodhia and Kish Greene and Lars Lowe Sjoesund and Lauren Usui and Laurent Sifre and Lena Heuermann and Leticia Lago and Lilly McNealus and Livio Baldini Soares and Logan Kilpatrick and Lucas Dixon and Luciano Martins and Machel Reid and Manvinder Singh and Mark Iverson and Martin Görner and Mat Velloso and Mateo Wirth and Matt Davidow and Matt Miller and Matthew Rahtz and Matthew Watson and Meg Risdal and Mehran Kazemi and Michael Moynihan and Ming Zhang and Minsuk Kahng and Minwoo Park and Mofi Rahman and Mohit Khatwani and Natalie Dao and Nenshad Bardoliwalla and Nesh Devanathan and Neta Dumai and Nilay Chauhan and Oscar Wahltinez and Pankil Botarda and Parker Barnes and Paul Barham and Paul Michel and Pengchong Jin and Petko Georgiev and Phil Culliton and Pradeep Kuppala and Ramona Comanescu and Ramona Merhej and Reena Jana and Reza Ardeshir Rokni and Rishabh Agarwal and Ryan Mullins and Samaneh Saadat and Sara Mc Carthy and Sarah Perrin and Sébastien M. R. Arnold and Sebastian Krause and Shengyang Dai and Shruti Garg and Shruti Sheth and Sue Ronstrom and Susan Chan and Timothy Jordan and Ting Yu and Tom Eccles and Tom Hennigan and Tomas Kocisky and Tulsee Doshi and Vihan Jain and Vikas Yadav and Vilobh Meshram and Vishal Dharmadhikari and Warren Barkley and Wei Wei and Wenming Ye and Woohyun Han and Woosuk Kwon and Xiang Xu and Zhe Shen and Zhitao Gong and Zichuan Wei and Victor Cotruta and Phoebe Kirk and Anand Rao and Minh Giang and Ludovic Peran and Tris Warkentin and Eli Collins and Joelle Barral and Zoubin Ghahramani and Raia Hadsell and D. Sculley and Jeanine Banks and Anca Dragan and Slav Petrov and Oriol Vinyals and Jeff Dean and Demis Hassabis and Koray Kavukcuoglu and Clement Farabet and Elena Buchatskaya and Sebastian Borgeaud and Noah Fiedel and Armand Joulin and Kathleen Kenealy and Robert Dadashi and Alek Andreev},
      year={2024},
      journal={ArXiv},
      url={https://arxiv.org/abs/2408.00118}, 
}

@article{ferrando2024primerinnerworkingstransformerbased,
      title={A Primer on the Inner Workings of Transformer-based Language Models}, 
      author={Javier Ferrando and Gabriele Sarti and Arianna Bisazza and Marta R. Costa-jussà},
      year={2024},
      eprint={2405.00208},
      journal={ArXiv},
      url={https://arxiv.org/abs/2405.00208}, 
}

@misc{zhang2024largemultimodalmodelsinterpret,
      title={Large Multi-modal Models Can Interpret Features in Large Multi-modal Models}, 
      author={Kaichen Zhang and Yifei Shen and Bo Li and Ziwei Liu},
      year={2024},
      eprint={2411.14982},
      archivePrefix={arXiv},
      primaryClass={cs.CV},
      url={https://arxiv.org/abs/2411.14982}, 
}

@misc{stevens2025sparseautoencodersscientificallyrigorous,
      title={Sparse Autoencoders for Scientifically Rigorous Interpretation of Vision Models}, 
      author={Samuel Stevens and Wei-Lun Chao and Tanya Berger-Wolf and Yu Su},
      year={2025},
      eprint={2502.06755},
      archivePrefix={arXiv},
      primaryClass={cs.CV},
      url={https://arxiv.org/abs/2502.06755}, 
}

@misc{xu2025decipheringfunctionsneuronsvisionlanguage,
      title={Deciphering Functions of Neurons in Vision-Language Models}, 
      author={Jiaqi Xu and Cuiling Lan and Xuejin Chen and Yan Lu},
      year={2025},
      eprint={2502.18485},
      archivePrefix={arXiv},
      primaryClass={q-bio.NC},
      url={https://arxiv.org/abs/2502.18485}, 
}

@misc{gurarieh2025enhancingautomatedinterpretabilityoutputcentric,
      title={Enhancing Automated Interpretability with Output-Centric Feature Descriptions}, 
      author={Yoav Gur-Arieh and Roy Mayan and Chen Agassy and Atticus Geiger and Mor Geva},
      year={2025},
      eprint={2501.08319},
      archivePrefix={arXiv},
      primaryClass={cs.CL},
      url={https://arxiv.org/abs/2501.08319}, 
}

@article{self_explaining_saes,
    author = {Dmitrii Kharlapenko and Stepan Shabalin and Neel Nanda and Arthur Conmy},
    title = {Self-explaining SAE features},
    journal={LessWrong},
    URL = {https://www.lesswrong.com/posts/8ev6coxChSWcxCDy8/self-explaining-sae-features},
    year = {2024}
}

@INPROCEEDINGS{imagenet,
  author={Deng, Jia and Dong, Wei and Socher, Richard and Li, Li-Jia and Kai Li and Li Fei-Fei},
  booktitle={2009 IEEE Conference on Computer Vision and Pattern Recognition}, 
  title={ImageNet: A large-scale hierarchical image database}, 
  year={2009},
  volume={},
  number={},
  pages={248-255},
  doi={10.1109/CVPR.2009.5206848}}

@misc{paulo2024automaticallyinterpretingmillionsfeatures,
      title={Automatically Interpreting Millions of Features in Large Language Models}, 
      author={Gonçalo Paulo and Alex Mallen and Caden Juang and Nora Belrose},
      year={2024},
      eprint={2410.13928},
      archivePrefix={arXiv},
      primaryClass={cs.LG},
      url={https://arxiv.org/abs/2410.13928}, 
}

@misc{choi2024automatic,
  author       = {Choi, Dami and Huang, Vincent and Meng, Kevin and Johnson, Daniel D and Steinhardt, Jacob and Schwettmann, Sarah},
  title        = {Scaling Automatic Neuron Description},
  year         = {2024},
  month        = {October},
  day          = {23},
  howpublished = {\url{https://transluce.org/neuron-descriptions}}
}

@misc{gemmateam2025gemma3technicalreport,
      title={Gemma 3 Technical Report}, 
      author={Gemma Team and Aishwarya Kamath and Johan Ferret and Shreya Pathak and Nino Vieillard and Ramona Merhej and Sarah Perrin and Tatiana Matejovicova and Alexandre Ramé and Morgane Rivière and Louis Rouillard and Thomas Mesnard and Geoffrey Cideron and Jean-bastien Grill and Sabela Ramos and Edouard Yvinec and Michelle Casbon and Etienne Pot and Ivo Penchev and Gaël Liu and Francesco Visin and Kathleen Kenealy and Lucas Beyer and Xiaohai Zhai and Anton Tsitsulin and Robert Busa-Fekete and Alex Feng and Noveen Sachdeva and Benjamin Coleman and Yi Gao and Basil Mustafa and Iain Barr and Emilio Parisotto and David Tian and Matan Eyal and Colin Cherry and Jan-Thorsten Peter and Danila Sinopalnikov and Surya Bhupatiraju and Rishabh Agarwal and Mehran Kazemi and Dan Malkin and Ravin Kumar and David Vilar and Idan Brusilovsky and Jiaming Luo and Andreas Steiner and Abe Friesen and Abhanshu Sharma and Abheesht Sharma and Adi Mayrav Gilady and Adrian Goedeckemeyer and Alaa Saade and Alex Feng and Alexander Kolesnikov and Alexei Bendebury and Alvin Abdagic and Amit Vadi and András György and André Susano Pinto and Anil Das and Ankur Bapna and Antoine Miech and Antoine Yang and Antonia Paterson and Ashish Shenoy and Ayan Chakrabarti and Bilal Piot and Bo Wu and Bobak Shahriari and Bryce Petrini and Charlie Chen and Charline Le Lan and Christopher A. Choquette-Choo and CJ Carey and Cormac Brick and Daniel Deutsch and Danielle Eisenbud and Dee Cattle and Derek Cheng and Dimitris Paparas and Divyashree Shivakumar Sreepathihalli and Doug Reid and Dustin Tran and Dustin Zelle and Eric Noland and Erwin Huizenga and Eugene Kharitonov and Frederick Liu and Gagik Amirkhanyan and Glenn Cameron and Hadi Hashemi and Hanna Klimczak-Plucińska and Harman Singh and Harsh Mehta and Harshal Tushar Lehri and Hussein Hazimeh and Ian Ballantyne and Idan Szpektor and Ivan Nardini and Jean Pouget-Abadie and Jetha Chan and Joe Stanton and John Wieting and Jonathan Lai and Jordi Orbay and Joseph Fernandez and Josh Newlan and Ju-yeong Ji and Jyotinder Singh and Kat Black and Kathy Yu and Kevin Hui and Kiran Vodrahalli and Klaus Greff and Linhai Qiu and Marcella Valentine and Marina Coelho and Marvin Ritter and Matt Hoffman and Matthew Watson and Mayank Chaturvedi and Michael Moynihan and Min Ma and Nabila Babar and Natasha Noy and Nathan Byrd and Nick Roy and Nikola Momchev and Nilay Chauhan and Noveen Sachdeva and Oskar Bunyan and Pankil Botarda and Paul Caron and Paul Kishan Rubenstein and Phil Culliton and Philipp Schmid and Pier Giuseppe Sessa and Pingmei Xu and Piotr Stanczyk and Pouya Tafti and Rakesh Shivanna and Renjie Wu and Renke Pan and Reza Rokni and Rob Willoughby and Rohith Vallu and Ryan Mullins and Sammy Jerome and Sara Smoot and Sertan Girgin and Shariq Iqbal and Shashir Reddy and Shruti Sheth and Siim Põder and Sijal Bhatnagar and Sindhu Raghuram Panyam and Sivan Eiger and Susan Zhang and Tianqi Liu and Trevor Yacovone and Tyler Liechty and Uday Kalra and Utku Evci and Vedant Misra and Vincent Roseberry and Vlad Feinberg and Vlad Kolesnikov and Woohyun Han and Woosuk Kwon and Xi Chen and Yinlam Chow and Yuvein Zhu and Zichuan Wei and Zoltan Egyed and Victor Cotruta and Minh Giang and Phoebe Kirk and Anand Rao and Kat Black and Nabila Babar and Jessica Lo and Erica Moreira and Luiz Gustavo Martins and Omar Sanseviero and Lucas Gonzalez and Zach Gleicher and Tris Warkentin and Vahab Mirrokni and Evan Senter and Eli Collins and Joelle Barral and Zoubin Ghahramani and Raia Hadsell and Yossi Matias and D. Sculley and Slav Petrov and Noah Fiedel and Noam Shazeer and Oriol Vinyals and Jeff Dean and Demis Hassabis and Koray Kavukcuoglu and Clement Farabet and Elena Buchatskaya and Jean-Baptiste Alayrac and Rohan Anil and Dmitry and Lepikhin and Sebastian Borgeaud and Olivier Bachem and Armand Joulin and Alek Andreev and Cassidy Hardin and Robert Dadashi and Léonard Hussenot},
      year={2025},
      eprint={2503.19786},
      archivePrefix={arXiv},
      primaryClass={cs.CL},
      url={https://arxiv.org/abs/2503.19786}, 
}

@misc{bai2025qwen25vltechnicalreport,
      title={Qwen2.5-VL Technical Report}, 
      author={Shuai Bai and Keqin Chen and Xuejing Liu and Jialin Wang and Wenbin Ge and Sibo Song and Kai Dang and Peng Wang and Shijie Wang and Jun Tang and Humen Zhong and Yuanzhi Zhu and Mingkun Yang and Zhaohai Li and Jianqiang Wan and Pengfei Wang and Wei Ding and Zheren Fu and Yiheng Xu and Jiabo Ye and Xi Zhang and Tianbao Xie and Zesen Cheng and Hang Zhang and Zhibo Yang and Haiyang Xu and Junyang Lin},
      year={2025},
      eprint={2502.13923},
      archivePrefix={arXiv},
      primaryClass={cs.CV},
      url={https://arxiv.org/abs/2502.13923}, 
}

@inproceedings{
ravi2025sam,
title={{SAM} 2: Segment Anything in Images and Videos},
author={Nikhila Ravi and Valentin Gabeur and Yuan-Ting Hu and Ronghang Hu and Chaitanya Ryali and Tengyu Ma and Haitham Khedr and Roman R{\"a}dle and Chloe Rolland and Laura Gustafson and Eric Mintun and Junting Pan and Kalyan Vasudev Alwala and Nicolas Carion and Chao-Yuan Wu and Ross Girshick and Piotr Dollar and Christoph Feichtenhofer},
booktitle={The Thirteenth International Conference on Learning Representations},
year={2025},
url={https://openreview.net/forum?id=Ha6RTeWMd0}
}

@misc{thasarathan2025universalsparseautoencodersinterpretable,
      title={Universal Sparse Autoencoders: Interpretable Cross-Model Concept Alignment}, 
      author={Harrish Thasarathan and Julian Forsyth and Thomas Fel and Matthew Kowal and Konstantinos Derpanis},
      year={2025},
      eprint={2502.03714},
      archivePrefix={arXiv},
      primaryClass={cs.CV},
      url={https://arxiv.org/abs/2502.03714}, 
}

@misc{gandelsman2023interpreting,
      title={Interpreting CLIP's Image Representation via Text-Based Decomposition}, 
      author={Yossi Gandelsman and Alexei A. Efros and Jacob Steinhardt},
      year={2023},
      eprint={2310.05916},
      archivePrefix={arXiv},
      primaryClass={cs.CV}
}

@misc{bhalla2024interpretingclipsparselinear,
      title={Interpreting CLIP with Sparse Linear Concept Embeddings (SpLiCE)}, 
      author={Usha Bhalla and Alex Oesterling and Suraj Srinivas and Flavio P. Calmon and Himabindu Lakkaraju},
      year={2024},
      eprint={2402.10376},
      archivePrefix={arXiv},
      primaryClass={cs.LG},
      url={https://arxiv.org/abs/2402.10376}, 
}

@InProceedings{10.1007/978-3-031-72980-5_26,
author="Rao, Sukrut
and Mahajan, Sweta
and B{\"o}hle, Moritz
and Schiele, Bernt",
editor="Leonardis, Ale{\v{s}}
and Ricci, Elisa
and Roth, Stefan
and Russakovsky, Olga
and Sattler, Torsten
and Varol, G{\"u}l",
title="Discover-then-Name: Task-Agnostic Concept Bottlenecks via Automated Concept Discovery",
booktitle="Computer Vision -- ECCV 2024",
year="2024",
publisher="Springer Nature Switzerland",
address="Cham",
pages="444--461",
isbn="978-3-031-72980-5"
}

@inproceedings{DBLP:conf/iclr/RaviGHHR0KRRGMP25,
  author       = {Nikhila Ravi and
                  Valentin Gabeur and
                  Yuan{-}Ting Hu and
                  Ronghang Hu and
                  Chaitanya Ryali and
                  Tengyu Ma and
                  Haitham Khedr and
                  Roman R{\"{a}}dle and
                  Chlo{\'{e}} Rolland and
                  Laura Gustafson and
                  Eric Mintun and
                  Junting Pan and
                  Kalyan Vasudev Alwala and
                  Nicolas Carion and
                  Chao{-}Yuan Wu and
                  Ross B. Girshick and
                  Piotr Doll{\'{a}}r and
                  Christoph Feichtenhofer},
  title        = {{SAM} 2: Segment Anything in Images and Videos},
  booktitle    = {The Thirteenth International Conference on Learning Representations,
                  {ICLR} 2025, Singapore, April 24-28, 2025},
  publisher    = {OpenReview.net},
  year         = {2025},
  url          = {https://openreview.net/forum?id=Ha6RTeWMd0},
  bibsource    = {dblp computer science bibliography, https://dblp.org}
}

@inproceedings{radford2021learning,
  title={Learning transferable visual models from natural language supervision},
  author={Radford, Alec and Kim, Jong Wook and Hallacy, Chris and Ramesh, Aditya and Goh, Gabriel and Agarwal, Sandhini and Sastry, Girish and Askell, Amanda and Mishkin, Pamela and Clark, Jack and others},
  booktitle={International conference on machine learning},
  pages={8748--8763},
  year={2021},
  organization={PmLR}
}

@inproceedings{reimers-gurevych-2019-sentence,
    title = "Sentence-{BERT}: Sentence Embeddings using {S}iamese {BERT}-Networks",
    author = "Reimers, Nils  and
      Gurevych, Iryna",
    editor = "Inui, Kentaro  and
      Jiang, Jing  and
      Ng, Vincent  and
      Wan, Xiaojun",
    booktitle = "Proceedings of the 2019 Conference on Empirical Methods in Natural Language Processing and the 9th International Joint Conference on Natural Language Processing (EMNLP-IJCNLP)",
    month = nov,
    year = "2019",
    address = "Hong Kong, China",
    publisher = "Association for Computational Linguistics",
    url = "https://aclanthology.org/D19-1410/",
    doi = "10.18653/v1/D19-1410",
    pages = "3982--3992"
}

@article{mcinnes2018umap-software,
  title={UMAP: Uniform Manifold Approximation and Projection},
  author={McInnes, Leland and Healy, John and Saul, Nathaniel and Grossberger, Lukas},
  journal={The Journal of Open Source Software},
  volume={3},
  number={29},
  pages={861},
  year={2018}
}

@misc{zhai2023sigmoidlosslanguageimage,
      title={Sigmoid Loss for Language Image Pre-Training}, 
      author={Xiaohua Zhai and Basil Mustafa and Alexander Kolesnikov and Lucas Beyer},
      year={2023},
      eprint={2303.15343},
      archivePrefix={arXiv},
      primaryClass={cs.CV},
      url={https://arxiv.org/abs/2303.15343}, 
}

@article{
oquab2024dinov,
title={{DINO}v2: Learning Robust Visual Features without Supervision},
author={Maxime Oquab and Timoth{\'e}e Darcet and Th{\'e}o Moutakanni and Huy V. Vo and Marc Szafraniec and Vasil Khalidov and Pierre Fernandez and Daniel HAZIZA and Francisco Massa and Alaaeldin El-Nouby and Mido Assran and Nicolas Ballas and Wojciech Galuba and Russell Howes and Po-Yao Huang and Shang-Wen Li and Ishan Misra and Michael Rabbat and Vasu Sharma and Gabriel Synnaeve and Hu Xu and Herve Jegou and Julien Mairal and Patrick Labatut and Armand Joulin and Piotr Bojanowski},
journal={Transactions on Machine Learning Research},
issn={2835-8856},
year={2024},
url={https://openreview.net/forum?id=a68SUt6zFt},
note={Featured Certification}
}

@misc{Joseph2024,
  author = {Joseph, Sonia},
  title = {Multimodal interpretability in 2024},
  year = {2024},
  howpublished = {\url{https://www.soniajoseph.ai/multimodal-interpretability-in-2024/}},
}

@inproceedings{
lim2025sparse,
title={Sparse autoencoders reveal selective remapping of visual concepts during adaptation},
author={Hyesu Lim and Jinho Choi and Jaegul Choo and Steffen Schneider},
booktitle={The Thirteenth International Conference on Learning Representations},
year={2025},
url={https://openreview.net/forum?id=imT03YXlG2}
}

@misc{
bai2024describeanddissect,
title={Describe-and-Dissect: Interpreting Neurons in Vision Networks with Language Models},
author={Nicholas Bai and Rahul Ajay Iyer and Tuomas Oikarinen and Tsui-Wei Weng},
year={2024},
url={https://openreview.net/forum?id=Rnxam2SRgB}
}

@misc{joseph2025steeringclipsvisiontransformer,
      title={Steering CLIP's vision transformer with sparse autoencoders}, 
      author={Sonia Joseph and Praneet Suresh and Ethan Goldfarb and Lorenz Hufe and Yossi Gandelsman and Robert Graham and Danilo Bzdok and Wojciech Samek and Blake Aaron Richards},
      year={2025},
      eprint={2504.08729},
      archivePrefix={arXiv},
      primaryClass={cs.CV},
      url={https://arxiv.org/abs/2504.08729}, 
}

@misc{marks2024dictionary_learning,
   title = {dictionary\_learning},
   author = {Samuel Marks and Adam Karvonen and Aaron Mueller},
   year = {2024},
   howpublished = {\url{https://github.com/saprmarks/dictionary_learning}},
}

@inproceedings{wolf-etal-2020-transformers,
    title = "Transformers: State-of-the-Art Natural Language Processing",
    author = "Thomas Wolf and Lysandre Debut and Victor Sanh and Julien Chaumond and Clement Delangue and Anthony Moi and Pierric Cistac and Tim Rault and Rémi Louf and Morgan Funtowicz and Joe Davison and Sam Shleifer and Patrick von Platen and Clara Ma and Yacine Jernite and Julien Plu and Canwen Xu and Teven Le Scao and Sylvain Gugger and Mariama Drame and Quentin Lhoest and Alexander M. Rush",
    booktitle = "Proceedings of the 2020 Conference on Empirical Methods in Natural Language Processing: System Demonstrations",
    month = oct,
    year = "2020",
    address = "Online",
    publisher = "Association for Computational Linguistics",
    url = "https://www.aclweb.org/anthology/2020.emnlp-demos.6",
    pages = "38--45"
}

@misc{esser2024scalingrectifiedflowtransformers,
      title={Scaling Rectified Flow Transformers for High-Resolution Image Synthesis}, 
      author={Patrick Esser and Sumith Kulal and Andreas Blattmann and Rahim Entezari and Jonas Müller and Harry Saini and Yam Levi and Dominik Lorenz and Axel Sauer and Frederic Boesel and Dustin Podell and Tim Dockhorn and Zion English and Kyle Lacey and Alex Goodwin and Yannik Marek and Robin Rombach},
      year={2024},
      eprint={2403.03206},
      archivePrefix={arXiv},
      primaryClass={cs.CV},
      url={https://arxiv.org/abs/2403.03206}, 
}

@misc{
huben2024sparse,
title={Sparse Autoencoders Find Highly Interpretable Features in Language Models},
author={Robert Huben and Hoagy Cunningham and Logan Riggs Smith and Aidan Ewart and Lee Sharkey},
booktitle={The Twelfth International Conference on Learning Representations},
year={2024},
url={https://openreview.net/forum?id=F76bwRSLeK}
}

@article{lloyd1982least,
  title={Least squares quantization in PCM},
  author={Lloyd, Stuart P},
  journal={IEEE Transactions on Information Theory},
  volume={28},
  number={2},
  pages={129--137},
  year={1982},
  publisher={IEEE}
}

@misc{rajaram2025linesightlinearrepresentations,
      title={Line of Sight: On Linear Representations in VLLMs}, 
      author={Achyuta Rajaram and Sarah Schwettmann and Jacob Andreas and Arthur Conmy},
      year={2025},
      eprint={2506.04706},
      archivePrefix={arXiv},
      primaryClass={cs.CV},
      url={https://arxiv.org/abs/2506.04706}, 
}

@misc{venhoff2025visualrepresentationsmaplanguage,
      title={How Visual Representations Map to Language Feature Space in Multimodal LLMs}, 
      author={Constantin Venhoff and Ashkan Khakzar and Sonia Joseph and Philip Torr and Neel Nanda},
      year={2025},
      eprint={2506.11976},
      archivePrefix={arXiv},
      primaryClass={cs.CV},
      url={https://arxiv.org/abs/2506.11976}, 
}

@misc{shabalin2025interpretinglargetexttoimagediffusion,
      title={Interpreting Large Text-to-Image Diffusion Models with Dictionary Learning}, 
      author={Stepan Shabalin and Ayush Panda and Dmitrii Kharlapenko and Abdur Raheem Ali and Yixiong Hao and Arthur Conmy},
      year={2025},
      eprint={2505.24360},
      archivePrefix={arXiv},
      primaryClass={cs.LG},
      url={https://arxiv.org/abs/2505.24360}, 
}

@article{arora-etal-2018-linear,
    title = "Linear Algebraic Structure of Word Senses, with Applications to Polysemy",
    author = "Arora, Sanjeev  and
      Li, Yuanzhi  and
      Liang, Yingyu  and
      Ma, Tengyu  and
      Risteski, Andrej",
    editor = "Lee, Lillian  and
      Johnson, Mark  and
      Toutanova, Kristina  and
      Roark, Brian",
    journal = "Transactions of the Association for Computational Linguistics",
    volume = "6",
    year = "2018",
    address = "Cambridge, MA",
    publisher = "MIT Press",
    url = "https://aclanthology.org/Q18-1034/",
    doi = "10.1162/tacl_a_00034",
    pages = "483--495"
}

@misc{gorton2024missingcurvedetectorsinceptionv1,
      title={The Missing Curve Detectors of InceptionV1: Applying Sparse Autoencoders to InceptionV1 Early Vision}, 
      author={Liv Gorton},
      year={2024},
      eprint={2406.03662},
      archivePrefix={arXiv},
      primaryClass={cs.LG},
      url={https://arxiv.org/abs/2406.03662}, 
}

@misc{joseph2025prismaopensourcetoolkit,
      title={Prisma: An Open Source Toolkit for Mechanistic Interpretability in Vision and Video}, 
      author={Sonia Joseph and Praneet Suresh and Lorenz Hufe and Edward Stevinson and Robert Graham and Yash Vadi and Danilo Bzdok and Sebastian Lapuschkin and Lee Sharkey and Blake Aaron Richards},
      year={2025},
      eprint={2504.19475},
      archivePrefix={arXiv},
      primaryClass={cs.CV},
      url={https://arxiv.org/abs/2504.19475}, 
}

@misc{fel2026rabbithulltaskrelevantconcepts,
      title={Into the Rabbit Hull: From Task-Relevant Concepts in DINO to Minkowski Geometry}, 
      author={Thomas Fel and Binxu Wang and Michael A. Lepori and Matthew Kowal and Andrew Lee and Randall Balestriero and Sonia Joseph and Ekdeep S. Lubana and Talia Konkle and Demba Ba and Martin Wattenberg},
      year={2026},
      eprint={2510.08638},
      archivePrefix={arXiv},
      primaryClass={cs.CV},
      url={https://arxiv.org/abs/2510.08638}, 
}

@inproceedings{
surkov2025onestep,
title={One-Step is Enough: Sparse Autoencoders for Text-to-Image Diffusion Models},
author={Viacheslav Surkov and Chris Wendler and Antonio Mari and Mikhail Terekhov and Justin Deschenaux and Robert West and Caglar Gulcehre and David Bau},
booktitle={The Thirty-ninth Annual Conference on Neural Information Processing Systems},
year={2025},
url={https://openreview.net/forum?id=MBJJ9Wcpg9}
}

@misc{karvonen2026activationoraclestrainingevaluating,
      title={Activation Oracles: Training and Evaluating LLMs as General-Purpose Activation Explainers}, 
      author={Adam Karvonen and James Chua and Clément Dumas and Kit Fraser-Taliente and Subhash Kantamneni and Julian Minder and Euan Ong and Arnab Sen Sharma and Daniel Wen and Owain Evans and Samuel Marks},
      year={2026},
      eprint={2512.15674},
      archivePrefix={arXiv},
      primaryClass={cs.CL},
      url={https://arxiv.org/abs/2512.15674}, 
}

@inproceedings{
pan2026latentqa,
title={Latent{QA}: Teaching {LLM}s to Decode Activations Into Natural Language},
author={Alexander Pan and Lijie Chen and Jacob Steinhardt},
booktitle={The Fourteenth International Conference on Learning Representations},
year={2026},
url={https://openreview.net/forum?id=niUroX9EOd}
}

@misc{li2026traininglanguagemodelsexplain,
      title={Training Language Models to Explain Their Own Computations}, 
      author={Belinda Z. Li and Zifan Carl Guo and Vincent Huang and Jacob Steinhardt and Jacob Andreas},
      year={2026},
      eprint={2511.08579},
      archivePrefix={arXiv},
      primaryClass={cs.CL},
      url={https://arxiv.org/abs/2511.08579}, 
}
}
\appendix
\appendix

\onecolumn
\section{SAE Training Details}\label{apx:sae_training}
For training the SAEs, we used the \texttt{dictionary\_learning} library~\citep{marks2024dictionary_learning}. All SAEs were optimized using the Adam optimizer with a learning rate of $3 \times 10^{-4}$, $\beta_1 = 0.9$, and $\beta_2 = 0.99$. Training was conducted over a single epoch of the ImageNet training set (1.28M images) with a batch size of 8192. We enforced a sparsity constraint of 25 active features per patch position.

Model activations from \texttt{HuggingFace}~\citep{wolf-etal-2020-transformers} were cached on-the-fly during training. We maintained a buffer of 500 million activations, from which we randomly sampled. When the buffer was depleted to half capacity, it was refilled with new activations.

\begin{table*}[!h]
\caption{Explanation evaluation metrics for the later layer SAE of Gemma 3 vision encoder. Except for AUROC, mean scores are reported, and statistical significance is assessed pairwise between methods. A value is underlined if it is significantly higher (with $p<0.05$) than both other methods in the same column.}
\centering
\resizebox{0.87\textwidth}{!}{%
\begin{tabular}{c *{8}{c}}
\toprule
\multirow{2}{*}{\textbf{Explanation Method}} 
& \multicolumn{2}{c}{\textbf{IoU Score}} 
& \multicolumn{2}{c}{\textbf{AUROC}} 
& \multicolumn{2}{c}{\textbf{Synth. Act. Score}} 
& \multicolumn{2}{c}{\textbf{CLIP Score}} \\
\cmidrule(lr){2-3} \cmidrule(lr){4-5} \cmidrule(lr){6-7} \cmidrule(lr){8-9}
\multicolumn{1}{c}{} & Masks & Heatmaps & Masks & Heatmaps & Masks & Heatmaps & Masks & Heatmaps \\
\midrule
\textbf{Steering} 
& \multicolumn{2}{c}{\underline{0.204}} 
& \multicolumn{2}{c}{0.773} 
& \multicolumn{2}{c}{1.473} 
& \multicolumn{2}{c}{0.182} \\
\textbf{Top-k}    
& 0.194 & 0.186 
& 0.782 & 0.857 
& 1.453 & 1.609 
& 0.188 & 0.187 \\
\textbf{Steering-informed Top-k} 
& 0.196 & 0.183 
& 0.810 & 0.908
& \underline{1.691} & \underline{2.156}
& \underline{0.190} & 0.186 \\
\bottomrule
\end{tabular}%
}
\vspace{5pt}
\label{tab:gemma3_later_masks_heatmaps}
\end{table*}

\section{Later Layer Results}\label{apx:later_layer}
See~\Cref{tab:gemma3_later_masks_heatmaps}.

\section{Statistical Test Details}\label{apx:stats_tests}
To assess statistical significance across explanation methods, we conduct pairwise one-tailed tests for each evaluation metric and masking type. Since evaluation scores are not normally distributed, as verified via a Shapiro-Wilk test, we apply the nonparametric Mann-Whitney U test. An explanation method is considered statistically significant if it is stochastically greater than both alternatives (with $p<0.05$).

\section{\topk Explanation Evaluation Scores as a Function of Semantic Similarity Between \steering and \topk Explanations}\label{apx:iou_clip_sim}

\begin{figure}[!h]
\begin{centering}
    \centering
    \includegraphics[width=0.45\textwidth]{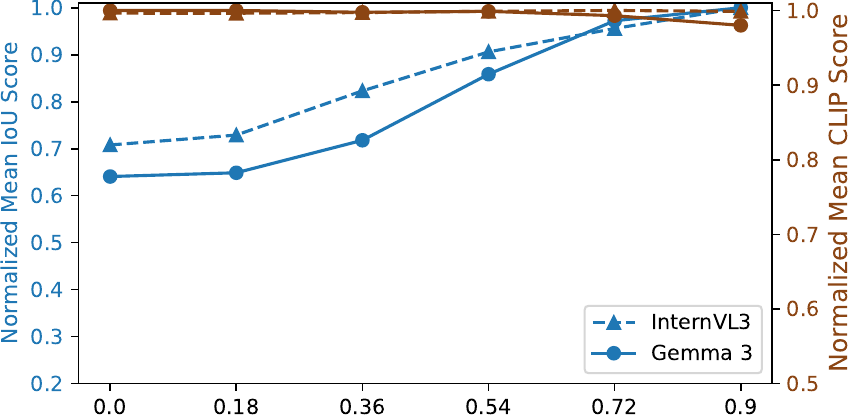}
    \label{fig:gemma_vs_intern_sim_threshold_mean_iou_clip_mean}
\caption{IoU Score and CLIP score values for \topk method as a function of the similarity with \steering explanations.}
\end{centering}
\end{figure}

\newpage
\section{Gemma 3 IoU and CLIP scores of \steering method as a function of the size of the LM $\text{m}_{\text{subj}}$}\label{apx:iou_clip_function_dec_size}
\begin{figure}[!h]
\begin{centering}
    \includegraphics[width=0.45\textwidth]{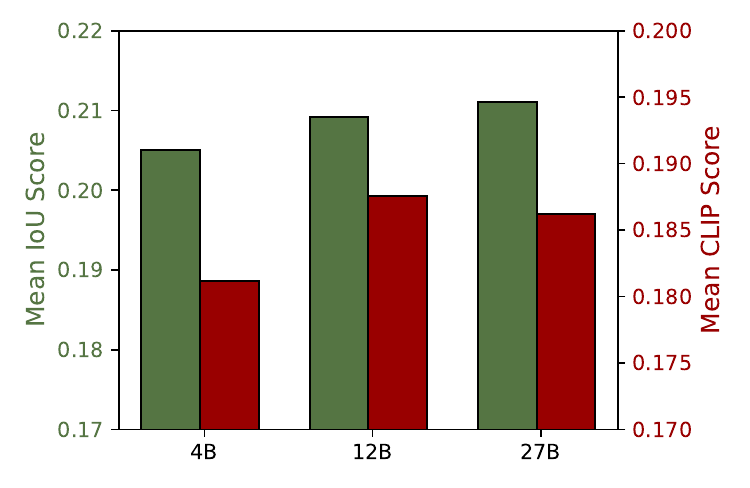}
    \caption{IoU Score and CLIP score values as a function of dataset size, for Masks Top-k method.}
    \label{fig:iou_clip_plot_mean_decoder_size}
\end{centering}
\end{figure}

\section{FLOPs Estimation} \label{apx:flops}
We compare the approximate Floating Point Operations (FLOPs) for generating explanations, using the estimate $2 \times \text{Parameters} \times \text{Tokens}$ for a model forward pass~\citep{kaplan2020scaling}.
Let:
\begin{itemize}
    \item $N_{\text{eval}} = |\mathcal{D}^{\text{eval}}|$: size of the evaluation image set.
    \item $P_{\text{sub}}$: parameters of the subject model $m_{\text{sub}}$ (also serving as $m_{\text{exp}}$).
    \item $P_{\text{SAE\_enc}} = d_{\text{model}} \cdot d_{\text{SAE}}$: parameters for the SAE.
    \item $T_{\text{img}}$: per image token representations (for $m_{\text{sub}}$ input, for SAE processing per image, and for the empty image $\tilde{I}$. E.g., 4096 for Gemma 3).
    \item $T_{\text{prompt}}$: token count for the textual prompt.
    \item $T_{\text{expl}}$: max tokens in the explanation.
    \item $k$: number of top images selected.
\end{itemize}

\paragraph{\topk Explanations.}
This method consists of two main computational stages:
\begin{enumerate}
    \item \textbf{Dataset Precomputation} (typically a one-time process to identify top-$k$ activating images for features): It involves processing all $N_{\text{eval}}$ images through $m_{\text{sub}}$, followed by SAE encoding for each representation using $\mW_{\text{enc}}$.
    $\text{FLOPs}_{\text{precompute}} \approx N_{\text{eval}} \cdot T_{\text{img}} \cdot 2 \cdot (P_{\text{sub}} + P_{\text{SAE\_enc}})$.
    Aggregation and sorting costs are generally minor in comparison.
    \item \textbf{Per-feature Explanation Generation}: The explainer model $m_{\text{sub}}$ is conditioned on the prompt and the $k$ selected images.
    $\text{FLOPs}_{\text{gen}} \approx 2 \cdot P_{\text{sub}} \cdot (T_{\text{prompt}} + k \cdot T_{\text{img}} + T_{\text{expl}})$.
\end{enumerate}
The total cost is dominated by $\text{FLOPs}_{\text{precompute}}$ when $N_{\text{eval}}$ is large.

\paragraph{\steering-based Explanations.}
This approach avoids the dataset precomputation. An explanation for each feature $i$ is generated via a single forward pass of $m_{\text{sub}}$ from an intervention using the pre-defined SAE feature direction $\mW_{\text{dec}}[i,:]$:
\begin{itemize}
    \item \textbf{Per-feature Explanation Generation}:
    $\text{FLOPs}_{\text{steer}} \approx 2 \cdot P_{\text{sub}} \cdot (T_{\text{prompt}} + T_{\text{img}} + T_{\text{expl}})$.
    The costs for retrieving the SAE feature direction and applying the intervention (vector operations) are also incurred, in addition to the forward pass captured by the formula above.
\end{itemize}

\paragraph{\stinformed Explanations.}
This method combines the dataset precomputation with an intervened generation step:
\begin{enumerate}
    \item \textbf{Dataset Precomputation}: This stage is identical to the corresponding stage in the \topk method, incurring $\text{FLOPs}_{\text{precompute}}$ as defined above.
    \item \textbf{Per-feature Explanation Generation}: Similar to standard \topk generation, but with an intervention. The computational cost for generation remains approximated by $\text{FLOPs}_{\text{gen}}$ as defined for \topk explanations. The costs for retrieving and applying the intervention are also incurred here, similar to the pure \textit{Steering-based} method. This method achieves the best results at a comparable cost.
\end{enumerate}

\newpage
\onecolumn
\section{Prompts}\label{apx:prompts}
\subsection{Explainer Prompts}
\begin{figure}[!h] 
    \centering 
    \begin{tcolorbox}[
        colback=gray!10!white,
        colframe=brown!70!black,
        title=Steering Prompt,
        fonttitle=\bfseries,
        boxrule=0.5pt,
        fontupper=\footnotesize,
        width=\textwidth 
    ]
\begin{verbatim}
You are given an image highlighting a visual or semantic element. This element may
range from a low-level visual feature to a high-level abstract concept. Your task is to
describe this element in a single, clear sentence. If the element is a high-level
abstract concept, describe it as such; otherwise, describe its visual patterns.
Favor a more general interpretation. Start the highlighted element description
with \"The highlighted element in the image is a\".

\end{verbatim}
\end{tcolorbox}
\caption{Prompt used for obtaining explanations for the \steering method. Gemma 3 outputs when given this prompt and a blank image (without steering) are reported in~\Cref{apx:prompt_only_behavior}.}
\label{fig:claim_extraction_system_prompt}
\end{figure}

\begin{figure}[!h] 
    \centering 
    \begin{tcolorbox}[
        colback=gray!10!white,
        colframe=brown!70!black,
        title=Top-k and Steering-informed Prompt (Masks),
        fonttitle=\bfseries,
        boxrule=0.5pt,
        fontupper=\footnotesize,
        width=\textwidth 
    ]
\begin{verbatim}
You are given set of images highlighting a visual or semantic element. The patches of
the images not showing the element are masked out, giving the impression of a
pixelated image. This element may range from a low-level visual feature to a high-level
abstract concept. Your task is to describe this element in a single, clear sentence.
If the element is a high-level abstract concept, describe it as such; otherwise,
describe its visual patterns. Favor a more general interpretation. Provide a single
description for the highlighted element appearing in all images, and please ignore the
pixelated effect of the mask when describing the element. Start the highlighted element
description with \"The highlighted element in the image is a\".

\end{verbatim}
\end{tcolorbox}
\caption{Prompt used for obtaining explanations for the \topk and \stinformed method with Masks.}
\label{fig:prompt_steering_informed_masks}
\end{figure}

\begin{figure}[!h] 
    \centering 
    \begin{tcolorbox}[
        colback=gray!10!white,
        colframe=brown!70!black,
        title=Top-k and Steering-informed Prompt (Heatmaps),
        fonttitle=\bfseries,
        boxrule=0.5pt,
        fontupper=\footnotesize,
        width=\textwidth 
    ]
\begin{verbatim}
You are given set of images highlighting a visual or semantic element. The patches of
the images showing the element are highlighted with a green heatmap. This element may
range from a low-level visual feature to a high-level abstract concept. Your task is
to describe this element in a single, clear sentence. If the element is a high-level
abstract concept, describe it as such; otherwise, describe its visual patterns. Favor
a more general interpretation. Provide a single description for the highlighted element
appearing in all images, and please ignore the overlayed green heatmap when describing
the element. Start the highlighted element description with \"The highlighted element
in the image is a\".

\end{verbatim}
\end{tcolorbox}
\caption{Prompt used for obtaining explanations for the \topk and \stinformed method with Heatmaps.}
\label{fig:prompt_steering_informed_heatmaps}
\end{figure}

\subsection{Prompt-only behavior}\label{apx:prompt_only_behavior}
To assess whether explanations may be driven by the prompt alone, we report model outputs when using the steering prompt with a blank image and no feature intervention. Smaller models correctly describe the absence of content (e.g., Gemma 3 4B: ``solid, uniformly white space, creating a blank canvas effect''; Gemma 3 12B: ``blank, white space''), while notably, the larger model produces a spurious description (Gemma 3 27B: ``stylized depiction of a bird in flight, characterized by its curved wings and streamlined body''), which might be explained by the unnatural input.

\newpage
\section{Models and Datasets}\label{apx:existing_assets}

We use the following assets in our work: 

\subsection*{Models}

\begin{table*}[h!]
\centering
\caption{The list of models used in this work.}
\label{tab:models}
\begin{tabular}{@{}ccc@{}} 
\toprule
\textbf{Model} & \textbf{Link} & \textbf{License} \\ \midrule
Gemma 3 \citep{gemmateam2025gemma3technicalreport} & \href{https://huggingface.co/google/gemma-3-27b-it}{Hugging Face (Google)} & Gemma Terms of Use \footnote{\url{https://ai.google.dev/gemma/terms}} \\
InternVL3-14B \citep{zhu2025internvl3} & \href{https://huggingface.co/OpenGVLab/InternVL3-14B}{Hugging Face (OpenGVLab)} & Apache 2.0 \\
CLIP \citep{radford2021learning} & \href{https://huggingface.co/openai/clip-vit-large-patch14}{Hugging Face (OpenAI)} & MIT License \\
SAM2 \citep{DBLP:conf/iclr/RaviGHHR0KRRGMP25} & \href{https://huggingface.co/facebook/sam2-hiera-large}{Hugging Face (Meta)} & Apache 2.0 \\ 
Stable Diffusion~\citep{esser2024scalingrectifiedflowtransformers} & \href{https://huggingface.co/stabilityai/stable-diffusion-3.5-medium}{Hugging Face (Stability AI)} & CreativeML OpenRAIL M license  \\
all-mpnet-base-v2~\citep{reimers-gurevych-2019-sentence} & \href{https://huggingface.co/sentence-transformers/all-mpnet-base-v2}{HuggingFace} & Apache 2.0 \\

\bottomrule
\end{tabular}
\end{table*}

\subsection*{Datasets}

\begin{table*}[h!]
\centering
\caption{The list of datasets used in this work.}
\label{tab:datasets}
\begin{tabular}{@{}ccc@{}} 
\toprule
\textbf{Dataset} & \textbf{Link} & \textbf{License} \\ \midrule
ImageNet \citep{imagenet} & \href{http://www.image-net.org}{Official Website} & Custom (Non-commercial) \\
\bottomrule
\end{tabular}
\end{table*}

\section{Compute Resources}\label{apx:compute}
All training and evaluation experiments were run on a single node of 4x NVIDIA Hopper H100 64GB GPUs. The demo website runs on a machine with 2x NVIDIA 4090 GPUs. Each Gemma 3 SAE training took approximately 6 hours on 1 GPU, and 3 hours for InternVL3.
\newpage
\section{Faithfulness Controls for Steering-Based Explanations}\label{apx:faithfulness_control}
\begin{table*}[!h]
\caption{Faithfulness controls for steering-based explanations on 1,000 SAE features (Gemma 3 vision encoder middle-layer). We compare true feature steering against random norm-matched directions and permuted feature directions. Mean scores shown. For Top-k-based explanations we used Masks.
}
\centering
\renewcommand{\arraystretch}{1.3}
\resizebox{0.75\textwidth}{!}{%
\begin{tabular}{c c c c c c}
\toprule
\textbf{Model} & \textbf{Explanation Method} & \textbf{IoU Score} & \textbf{AUROC} & \textbf{Synth. Act. Score} & \textbf{CLIP Score} \\
\midrule
\multirow{5}{*}{\raisebox{-2.5em}{\rotatebox{90}{Gemma 3}}}
& \textbf{Steering}                & 0.204 & 0.691 & 0.353 & 0.186 \\
& \textbf{Top-k}                   & 0.206 & 0.740 & 0.365 & 0.190 \\
& \textbf{Steering-informed Top-k} & 0.209 & 0.806 & 0.518 & 0.193 \\
& \textbf{Random-vector Steering}  & 0.149 & 0.439 & 0.008 & 0.179 \\
& \textbf{Random-perm Steering}    & 0.138 & 0.448 & 0.008 & 0.161 \\
\bottomrule
\end{tabular}%
}
\label{tab:mid_masks_eval_results}
\end{table*}

To assess whether steering-based explanations are feature-specific rather than driven by the prompt or generic activation effects, we introduce two control baselines: (i) random norm-matched directions, and (ii) random permutations of SAE feature directions. In both cases, the intervention strength is matched to the original steering setup.

We evaluate these controls on 1,000 SAE features from the middle layer of the Gemma 3 vision encoder, using Gemma 3 27B as the explainer language model.~\Cref{tab:mid_masks_eval_results} reports the results.

We observe that both control baselines perform substantially worse than true feature steering across all metrics. In particular, synthetic activation drops to near-zero and AUROC converges to near 0.5 (random-classifier level), lack causal alignment with the underlying feature activations. IoU and CLIP scores are also consistently lower.

\end{document}